\documentclass{article}

\usepackage{times}
\usepackage{graphicx} 
\usepackage{subfigure} 

\usepackage{natbib}

\usepackage{algorithm}
\usepackage{algorithmic}

\usepackage{hyperref}

\usepackage[accepted]{icml2016}

\icmltitlerunning{Deep Perceptual Similarity Metrics}

\usepackage{amsmath}
\usepackage{amssymb}
\usepackage{color}
\usepackage{multirow}
\usepackage{array}
\usepackage{caption}

\definecolor{dgreen}{rgb}{0,.7,0}
\definecolor{dyellow}{rgb}{.7,.7,0}
\definecolor{dred}{rgb}{.7,0,0}
\definecolor{dblue}{rgb}{0,0,0.7}

\definecolor{alexey}{rgb}{0.7,0,1}

\newcommand{\oR}{\mathbb{R}}

\newcommand{\conv}{\textsc{conv}}

\newcommand{\fc}{\textsc{fc}}

\newcommand{\ourapproach}{DeePSiM }
\newcommand{\inp}{\mathbf{x}}
\newcommand{\targ}{\mathbf{y}}
\newcommand{\img}{\mathbf{I}}
\newcommand{\feat}{\mathbf{\phi}}
\newcommand{\weights}{\mathbf{\theta}}
\newcommand{\recimg}{\widetilde{\img}}
\newcommand{\gen}{G_\weights}
\newcommand{\comp}{C}
\newcommand{\repres}{\Phi}
\newcommand{\discr}{D_\varphi}
\newcommand{\prior}{P}
\newcommand{\loss}{\mathcal{L}}
\newcommand{\ltwo}{SE }
\newcommand{\lone}{$\ell_1$ }
\newcommand{\featloss}{\loss_{feat}}
\newcommand{\discrloss}{\loss_{adv}}
\newcommand{\pixloss}{\loss_{img}}
\newcommand{\klloss}{\loss_{KL}}
\newcommand{\imgspace}{\oR^{W \times H \times C}}
\newcommand{\inspace}{\oR^I}
\newcommand{\featspace}{\oR^F}
\newcommand{\enc}{Enc}
\newcommand{\dec}{Dec}
\newcommand{\vaein}{x}
\newcommand{\vaerec}{\tilde{\vaein}}
\newcommand{\latent}{z}
\newcommand{\expect}{\mathbb{E}}
\newcommand{\kldiv}{D_{KL}}
\newcommand{\vaemean}{\mathbf{\mu}}
\newcommand{\vaesigma}{\mathbf{\sigma}}
\newcommand{\vaenoise}{\mathbf{\varepsilon}}

\graphicspath{{images/}}

\begin{document} 

\twocolumn[
\icmltitle{Generating Images with Perceptual Similarity Metrics based on Deep Networks}

\icmlauthor{Alexey Dosovitskiy}{dosovits@cs.uni-freiburg.de}
\icmladdress{University of Freiburg, Germany}
\icmlauthor{Thomas Brox}{brox@cs.uni-freiburg.de}
\icmladdress{University of Freiburg, Germany}

\icmlkeywords{convolutional networks, generative models, image generation}

\vskip 0.3in
]

\begin{abstract}
Image-generating machine learning models are typically trained with loss functions based on distance in the image space.
This often leads to over-smoothed results.
We propose a class of loss functions, which we call deep perceptual similarity metrics (DeePSiM), that mitigate this problem. 
Instead of computing distances in the image space, we compute distances between image features extracted by deep neural networks.
This metric better reflects perceptually similarity of images and thus leads to better results.
We show three applications: autoencoder training, a modification of a variational autoencoder, and inversion of deep convolutional networks.
In all cases, the generated images look sharp and resemble natural images.
\end{abstract}


\section{Introduction}
\begin{figure}[b]
\begin{center}
\setlength{\tabcolsep}{0.1cm}
\renewcommand{\arraystretch}{1}
\small{
  \begin{tabular}{c}
  Original \;\;\; Img loss \; Img + Adv\,  Img + Feat \quad\; Our\;\;\;\;  \\
  \includegraphics[width=0.9\linewidth]{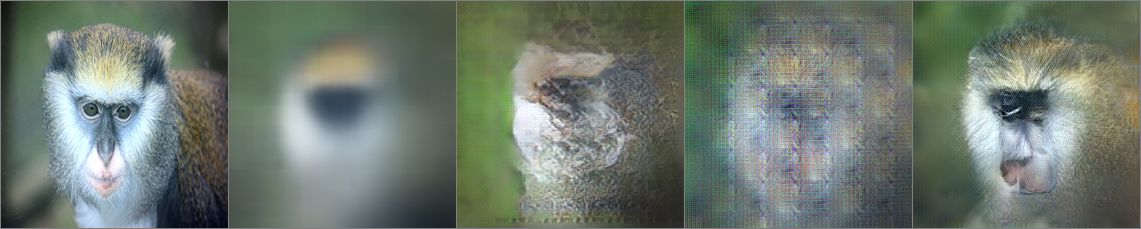}\\
  a) \quad\qquad\;\; b) \quad\qquad\; c) \quad\qquad\; d) \qquad\quad\;\; e)
   \end{tabular}}
   \vspace{-0.15cm}
\end{center}
   \caption{Reconstructions from layer \fc6 of AlexNet with different losses.}
\label{fig:ablation_intro}
\end{figure}

Recently there has been a surge of interest in training neural networks to generate images.
These are being used for a wide variety of applications: unsupervised and semi-supervised learning, generative models, analysis of learned representations, analysis by synthesis, learning of 3D representations, future prediction in videos.
Nevertheless, there is little work on studying loss functions which are appropriate for the image generation task.
Typically used squared Euclidean distance between images often yields blurry results, see Fig.\ref{fig:ablation_intro}b.
This is especially the case when there is inherent uncertainty in the prediction.
For example, suppose we aim to reconstruct an image from its feature representation.
The precise location of all details may not be preserved in the features.
A loss in image space leads to averaging all likely locations of details, and hence the reconstruction looks blurry.

However, exact locations of all fine details are not important for perceptual similarity of images.
But the distribution of these details plays a key role. 
Our main insight is that invariance to irrelevant transformations and sensitivity to local image statistics can be achieved by measuring distances in a suitable feature space. 
In fact, convolutional networks provide a feature representation with desirable properties.
They are invariant to small smooth deformations, but sensitive to perceptually important image properties, for example sharp edges and textures.

Using a distance in feature space alone, however, does not yet yield a good loss function; see Fig.~\ref{fig:ablation_intro}d.
Since feature representations are typically contractive, many images, including non-natural ones, get mapped to the same feature vector.
Hence, we must introduce a natural image prior.
To this end, we build upon adversarial training as proposed by~\citet{Goodfellow_NIPS2014}.
We train a discriminator network to distinguish the output of the generator from real images.
The objective of the generator is to trick the discriminator, i.e., to generate images that the discriminator cannot distinguish from real ones. 
This yields a natural image prior that selects from all potential generator outputs the most realistic one. 
A combination of similarity in an appropriate feature space with adversarial training allows to obtain the best results; see Fig.~\ref{fig:ablation_intro}e.


We show three example applications: image compression with an autoencoder, a generative model based on a variational autoencoder, and inversion of the AlexNet convolutional network. 
We demonstrate that an autoencoder with \ourapproach loss can compress images  while preserving information about fine structures.
On the generative modeling side, we show that a version of a variational autoencoder trained with the new loss produces images with realistic image statistics. 
Finally, reconstructions obtained with our method from high-level activations of AlexNet are dramatically better than with existing approaches. They demonstrate that even the predicted class probabilities contain rich texture, color, and position information.


\section{Related work}
There is a long history of neural network based models for image generation.
A prominent class of probabilistic models of images are restricted Boltzmann machines~\citep{Hinton_1986, Smolensky_1986, Hinton_Science2006} and their deep variants~\citep{Hinton_NC2006,Salakhutdinov_2009,Lee_ICML2009}.
Autoencoders~\citep{Hinton_Science2006, Vincent_ICML2008} have been widely used for unsupervised learning and generative modeling, too.
Recently, stochastic neural networks~\citep{Bengio_ICML2014, Kingma_NIPS2014, Gregor_ICML2015} have become popular, and deterministic networks are being used for image generation tasks~\citep{Dosovitskiy_CVPR2015}.
In all these models, loss is measured in the image space.
By combining convolutions and un-pooling (upsampling) layers~\citep{Lee_ICML2009, Goodfellow_NIPS2014, Dosovitskiy_CVPR2015} these models can be applied to large images.

There is a large body of work on assessing the perceptual similarity of images.
Some prominent examples are the visible differences predictor ~\citep{Daly_1993}, the spatio-temporal model for moving picture quality assessment ~\citep{Lambrecht_1996}, and the perceptual distortion metric of ~\citet{Winkler_1998}.
The most popular perceptual image similarity metric is the structural similarity metric (SSIM)~\citep{Wang_2004}, which compares the local statistics of image patches.
We are not aware of any work making use of similarity metrics for machine learning, except a recent pre-print of~\citet{Ridgeway_arxiv15}.
They train autoencoders by directly maximizing the SSIM similarity of images.
This resembles in spirit what we do, but technically is very different. 
While psychophysical experiments go out of scope of this paper, we believe that deep learned feature representations have better potential than shallow hand-designed SSIM.

Generative adversarial networks (GANs) have been proposed by~\citet{Goodfellow_NIPS2014}.
In theory, this training procedure can lead to a generator that perfectly models the data distribution.
Practically, training GANs is difficult and often leads to oscillatory behavior, divergence, or modeling only part of the data distribution.
Recently, several modifications have been proposed that make GAN training more stable.
\citet{Denton_NIPS2015} employ a multi-scale approach, gradually generating higher resolution images.
\citet{Radford_arxiv2015} make use of a convolutional-deconvolutional architecture and batch normalization.

GANs can be trained conditionally by feeding the conditioning variable to both the discriminator and the generator~\citep{Mirza_2014}.
Usually this conditioning variable is a one-hot encoding of the object class in the input image.
Such GANs learn to generate images of objects from a given class.
Recently~\citet{Mathieu_arxiv2015} used GANs for predicting future frames in videos by conditioning on previous frames.
Our approach looks similar to a conditional GAN. 
However, in a GAN there is no loss directly comparing the generated image to some ground truth.
We found that the feature loss introduced in the present paper is essential to train on complicated tasks such as feature inversion.

Most related is concurrent work of~\citet{Larsen_arxiv2015}.
The general idea is the same~--- to measure the similarity not in the image space, but rather in a feature space.
They also use adversarial training to improve the realism of the generated images.
However, \citet{Larsen_arxiv2015} only apply this approach to a variational autoencoder trained on images of faces, and measure the similarity between features extracted from the discriminator.
Our approach is much more general, we apply it to various natural images, and we demonstrate three different applications. 


\section{Model}
Suppose we are given a supervised learning task and a training set of input-target pairs $\{\inp_i,\, \targ_i\}$, $\inp_i \in \inspace$, $\targ_i \in \imgspace$ .
Inputs and outputs can be arbitrary vectors. In this work, we focus on targets that are images with an arbitrary number of channels. 

The aim is to learn the parameters $\weights$ of a differentiable generator function $\gen(\cdot) \colon \inspace \to \imgspace$ that optimally approximates the input-target dependency according to a loss function $\loss (\gen(\inp), \targ)$.
Typical choices are squared Euclidean (\ltwo\!\!) loss $\loss_2 (\gen(\inp), \targ) = || \gen(\inp) - \targ ||_2^2$ or \lone loss $\loss_1 (\gen(\inp),\targ) = ||\gen(\inp)-\targ||_1$.
As we demonstrate in this paper, these losses are suboptimal for some image generation tasks. 

\begin{figure}
\begin{center}
\includegraphics[width=0.85\linewidth]{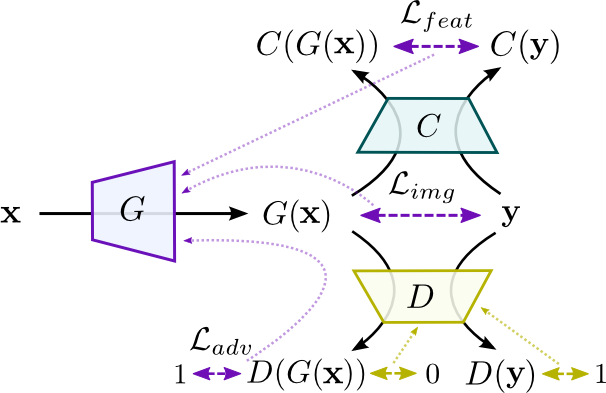}
   \caption{Schematic of our model. Black solid lines denote the forward pass. 
   Dashed lines with arrows on both ends are the losses. 
   Thin dashed lines denote the flow of gradients.}
   \label{fig:model}
\end{center}
\end{figure}

We propose a new class of losses, which we call DeePSiM.
These go beyond simple distances in image space and can capture complex and perceptually important properties of images.
These losses are weighted sums of three terms: feature loss $\featloss$, adversarial loss $\discrloss$, and pixel space loss $\pixloss$:
\begin{equation}
 \loss = \lambda_{feat}\, \featloss  + \lambda_{adv}\, \discrloss + \lambda_{img}\, \pixloss.
\end{equation}
They correspond to a network architecture, an overview of which is shown in Fig.~\ref{fig:model}. The architecture consists of three convolutional networks: the generator $G$ that implements the generator function, the discriminator $\discr$ that discriminates generated images from natural images, and the comparator $C$ that computes features from images.     

\textbf{Loss in feature space.} 
Given a differentiable comparator $\comp \colon \imgspace \to \featspace$, we define
\begin{equation}
 \featloss = \sum\limits_{i} || \comp(\gen(\inp_i)) - \comp(\targ_i) ||_2^2.
\end{equation}
$\comp$ may be fixed or may be trained; for example, it can be a part of the generator or the discriminator.

$\featloss$ alone does not provide a good loss for training.
It is known~\citep{Mahendran_CVPR2015} that optimizing just for similarity in the feature space typically leads to high-frequency artifacts.
This is because 
for each natural image there are many non-natural images mapped to the same feature vector
\footnote{This is unless the feature representation is specifically designed to map natural and non-natural images far apart, such as the one extracted from the discriminator of a GAN.}.
Therefore, a natural image prior is necessary to constrain the generated images to the manifold of natural images.

\textbf{Adversarial loss.} Instead of manually designing a prior, as in \citet{Mahendran_CVPR2015}, we learn it with an approach similar to Generative Adversarial Networks (GANs) of \citet{Goodfellow_NIPS2014}.
Namely, we introduce a discriminator $\discr$ which aims to discriminate the generated images from real ones, and which is trained concurrently with the generator $\gen$.
The generator is trained to ``trick'' the discriminator network into classifying the generated images as real.
Formally, the parameters $\varphi$ of the discriminator are trained by minimizing
\begin{equation} \label{eq:discrloss_discr}
 \mathcal{L}_{discr} = - \sum\limits_{i} \log (\discr(\targ_i)) + \log (1 - \discr(\gen(\inp_i))), 
\end{equation}
and the generator is trained to minimize
\begin{equation} \label{eq:discrloss_gen}
 \discrloss = - \sum\limits_{i} \log \discr(\gen(\inp_i)).
\end{equation}

\textbf{Loss in image space.}
Adversarial training is known to be unstable and sensitive to hyperparameters.
We found that adding a loss in the image space
\begin{equation}
 \pixloss = \sum\limits_{i} || \gen(\inp_i) - \targ_i ||_2^2.
\end{equation}
stabilizes training.

\subsection{Architectures}

\textbf{Generators.} 
We used several different generators in experiments. 
They are task-specific, so we describe these in corresponding sections below.
All tested generators make use of up-convolutional ('deconvolutional') layers, as in~\citet{Dosovitskiy_CVPR2015}.
An up-convolutional layer consists of up-sampling and a subsequent convolution.
In this paper we always up-sample by a factor of $2$ and a 'bed of nails' upsampling.

In all networks we use leaky ReLU nonlinearities, that is, $LReLU(x) = \max(x,0) + \alpha \min(x,0)$.
We used $\alpha = 0.3$ in our experiments.
All generators have linear output layers.

\textbf{Comparators.}
We experimented with four comparators:

1. AlexNet ~\citep{Krizhevsky_NIPS2012} is a network with $5$ convolutional and $2$ fully connected layers trained on image classification.

2. The network of \citet{Wang_ICCV2015} has the same architecture as AlexNet, but is trained using videos with triplet loss, which enforces frames of one video to be close in the feature space and frames from different videos to be far apart. We refer to this network as VideoNet.

3. AlexNet with random weights.

4. Exemplar-CNN ~\citep{Exemplar_PAMI2015} is a network with $3$ convolutional layers and $1$ fully connected layer trained on a surrogate task of discriminating between different image patches.

The exact layers used for comparison are specified in the experiments sections.

\textbf{Discriminator.} The architecture of the discriminator was nearly the same in all experiments.
The version used for the autoencoder experiments is shown in Table~\ref{tbl:discriminator_arch}.
The discriminator must ensure the local statistics of images to be natural.
Therefore after five convolutional layers with occasional stride we perform global average pooling.
The result is processed by two fully connected layers, followed by a $2$-way softmax.
We perform $50 \%$ dropout after the global average pooling layer and the first fully connected layer.

There are two modifications to this basic architecture.
First, when dealing with large ImageNet~\citep{imagenet} images we increase the stride in the first layer from $2$ to $4$.
Second, when training networks to invert AlexNet, we additionally feed the features to the discriminator.
We process them with two fully connected layers with $1024$ and $512$ units, respectively. 
Then we concatenate the result with the output of global average pooling.

\begin{table}
   \begin{center}
   \setlength{\tabcolsep}{0.15cm}
  \small{
  \begin{tabular}{|l|cccccccc|}
      \hline
      \small{Type}   & conv  & conv & conv  & conv  & conv  & pool  & fc    & fc   \\ \hline
      \small{InSize} & $64$  & $29$ & $25$  & $12$  & $10$  & $4$   & $-$   & $-$  \\
      \small{OutCh}  & $32$  & $64$ & $128$ & $256$ & $256$ & $256$ & $512$ & $2$  \\
      \small{Kernel}    & $7$   & $5$  & $3$   & $3$   & $3$   & $4$   & $-$   & $-$  \\
      \small{Stride}    & $2$   & $1$  & $2$   & $1$   & $2$   & $4$   & $-$   & $-$  \\
      \hline
    \end{tabular}}
  \end{center}
  \caption{Discriminator architecture. }
  \label{tbl:discriminator_arch}
\end{table}

\subsection{Training details}

We modified the \emph{caffe}~\citep{caffe} framework to train the networks.
For optimization we used Adam~\citep{Kingma_ICLR2015} with momentum $\beta_1=0.9$, $\beta_2=0.999$ and initial learning rate $0.0002$.
To prevent the discriminator from overfitting during adversarial training we temporarily stopped updating it if the ratio of $\mathcal{L}_{discr}$ and $\discrloss$ was below a certain threshold ($0.1$ in most experiments). 
We used batch size $64$ in all experiments. 
We trained for $500,000$-$1,000,000$ mini-batch iterations.

\section{Experiments}
We started with a simple proof-of-concept experiment showing how \ourapproach can be applied to training autoencoders.
Then we used the proposed loss function within the variational autoencoder (VAE) framework.
Finally, we applied the method to invert the representation learned by AlexNet and analyzed some properties of the method.

In quantitative comparisons we report normalized Euclidean error $||a-b||_2 / N$.
The normalization coefficient $N$ is the average of Euclidean distances between all pairs of different samples from the test set.
Therefore, the error of $100 \%$ means that the algorithm performs the same as randomly drawing a sample from the test set.

\subsection{Autoencoder} \label{sec:exp_ae}

Here the target of the generator coincides with its input (that is, $\targ = \inp$), and the task of the generator is to encode the input to a compressed hidden representation and then decode back the image. 
The architecture is shown in Table~\ref{tbl:ae_arch}.
All layers are convolutional or up-convolutional.
The hidden representation is an $8$-channel feature map $8$ times smaller than the input image.
We trained on the STL-10~\citep{Coates_AISTATS2010} unlabeled dataset which contains $100,000$ images $96 \times 96$ pixels.
To prevent overfitting we augmented the data by cropping random $64 \times 64$ patches during training.

We experimented with four loss functions: \ltwo and \lone in the image space, as well as \ourapproach with AlexNet \conv3 or Exemplar-CNN \conv3 as comparator.

Qualitative results are shown in Fig.~\ref{fig:ae_qualitative}, quantitative results in Table~\ref{tbl:ae_quantitative}.
While underperforming in terms of Euclidean loss, our approach can preserve more texture details, resulting in naturally looking non-blurry reconstructions.
Interestingly, AlexNet as comparator tends to corrupt fine details (petals of the flower, sails of the ship), perhaps because it has stride of $4$ in the first layer.
Exemplar-CNN as comparator does not preserve the exact color because it is explicitly trained to be invariant to color changes. 
We believe that with carefully selected or specifically trained comparators yet better results can be obtained.

We stress that lower Euclidean error does not mean better reconstruction.
For example, imagine a black-and-white striped "zebra" pattern.
A monotonous gray image will have twice smaller Euclidean error than the same pattern shifted by one stripe width.

\begin{table}
\begin{center}
   \setlength{\tabcolsep}{0.18cm}
  \small{
  \begin{tabular}{|l|cccccccc|}
      \hline
      \small{InSize} & $64$               & $32$ & $32$               & $16$  & $16$               & $8$   & $8$   & $8$  \\
      \small{OutCh}  & $32$               & $32$ & $64$               & $64$  & $128$              & $128$ & $64$  & $8$  \\
      \small{Kernel} & $5$                & $3$  & $5$                & $3$   & $3$                & $3$   & $3$   & $3$  \\
      \small{Stride} & $\downarrow \!2$   & $1$  & $\downarrow \!2$   & $1$   & $\downarrow \!2$   & $1$   & $1$   & $1$  \\
      \hline
    \end{tabular}}
\begin{center}
\end{center}
    \vspace{0.03cm}
    \small{
  \begin{tabular}{|l|cccccccc|}
      \hline
      \small{InSize} & $8$  & $8$   & $8$            & $16$  & $16$           & $32$  & $32$             & $64$ \\
      \small{OutCh}  & $64$ & $128$ & $64$           & $64$  & $32$           & $32$  & $16$             & $3$  \\
      \small{Kernel} & $3$  & $3$   & $4$            & $3$   & $4$            & $3$   & $4$              & $3$  \\
      \small{Stride} & $1$  & $1$   & $\uparrow \!2$ & $1$   & $\uparrow \!2$ & $1$   & $\uparrow \!2$   & $1$  \\
      \hline
    \end{tabular}}
\end{center}
  \caption{Autoencoder architecture. \textbf{Top:} encoder, \textbf{bottom:} decoder. All layers are convolutional or 'up-convolutional'.}
  \label{tbl:ae_arch}
\end{table}

\begin{figure}
\begin{center}
\setlength{\tabcolsep}{0.1cm}
\renewcommand{\arraystretch}{1}
  \begin{tabular}{cc}
  Image &
  \raisebox{-.5\height}{\includegraphics[width=0.8\linewidth]{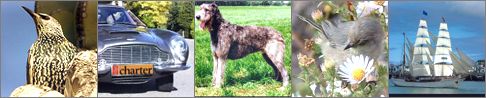}} \\
  AlexNet &
  \raisebox{-.5\height}{\includegraphics[width=0.8\linewidth]{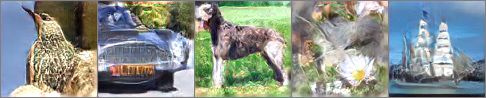}} \\ 
  Ex-CNN &
  \raisebox{-.5\height}{\includegraphics[width=0.8\linewidth]{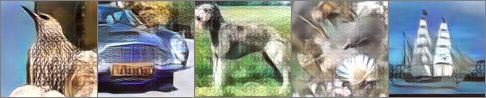}} \\ 
  \ltwo &
  \raisebox{-.5\height}{\includegraphics[width=0.8\linewidth]{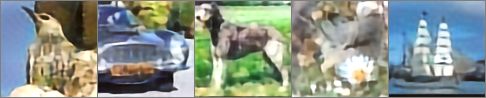}} \\ 
  \lone &
  \raisebox{-.5\height}{\includegraphics[width=0.8\linewidth]{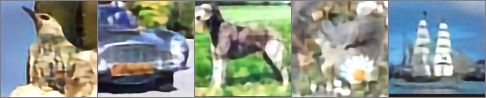}} \\ 
   \end{tabular}
\end{center}
   \caption{Autoencoder qualitative results. Best viewed on screen.}
\label{fig:ae_qualitative}
\end{figure}

\begin{table}
\begin{center}
\vspace{0.5cm}
  \begin{tabular}{c|c|c|c}
  \ltwo loss     & \lone loss     & Our-ExCNN      & Our-AlexNet     \\ \hline
  $15.3$         & $15.7$         & $19.8$         & $21.5$ 
   \end{tabular}
   \vspace{-0.2cm}
\end{center}
   \caption{Normalized Euclidean reconstruction error (in \%) of autoencoders trained with different loss functions.}
\label{tbl:ae_quantitative}
\vspace{-0.2cm}
\end{table}

\begin{table}
\begin{center}
  \begin{tabular}{c|c|c|c}
  \ltwo loss     & \lone loss     & Our-ExCNN      & Our-AlexNet     \\ \hline
  $34.6 \pm 0.6$ & $35.7 \pm 0.4$ & $50.1 \pm 0.5$ & $52.3 \pm 0.6$ 
   \end{tabular}
   \vspace{-0.25cm}
\end{center}
   \caption{Classification accuracy (in \%) on STL with autoencoder features learned with different loss functions.}
\label{tbl:ae_classification}
\vspace{-0.25cm}
\end{table}

\textbf{Classification.}
Reconstruction-based models are commonly used for unsupervised feature learning.
We checked if our loss functions lead to learning more meaningful representations than usual \lone and \ltwo losses.
To this end, we trained linear SVMs on the $8$-channel hidden representations extracted by autoencoders trained with different losses.
We are just interested in relative performance and, thus, do not compare to the state of the art.
We trained on $10$ folds of the STL-10 training set and tested on the test set.

The results are shown in Table~\ref{tbl:ae_classification}.
As expected, the features learned with \ourapproach perform significantly better, indicating that they contain more semantically meaningful information.
This suggests that other losses than standard \lone and \ltwo may be useful for unsupervised learning. Note that the Exemplar-CNN comparator is trained in an unsupervised way.

\begin{figure}[b!]
\begin{center}
\vspace{-0.25cm}
\begin{tabular}{c}
\includegraphics[width=0.95\linewidth]{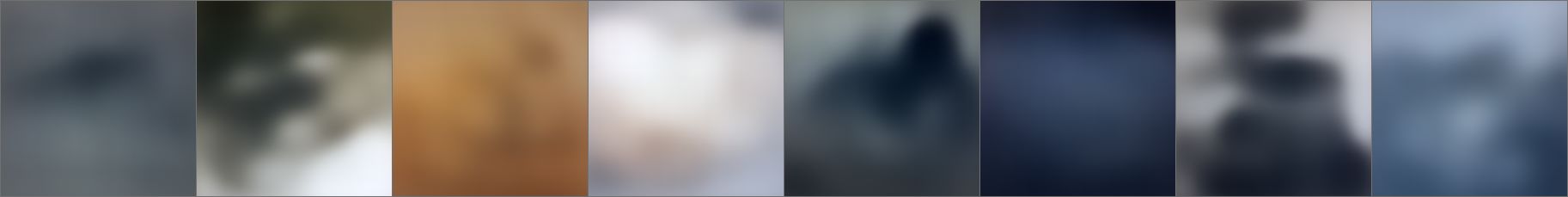}\\
\includegraphics[width=0.95\linewidth]{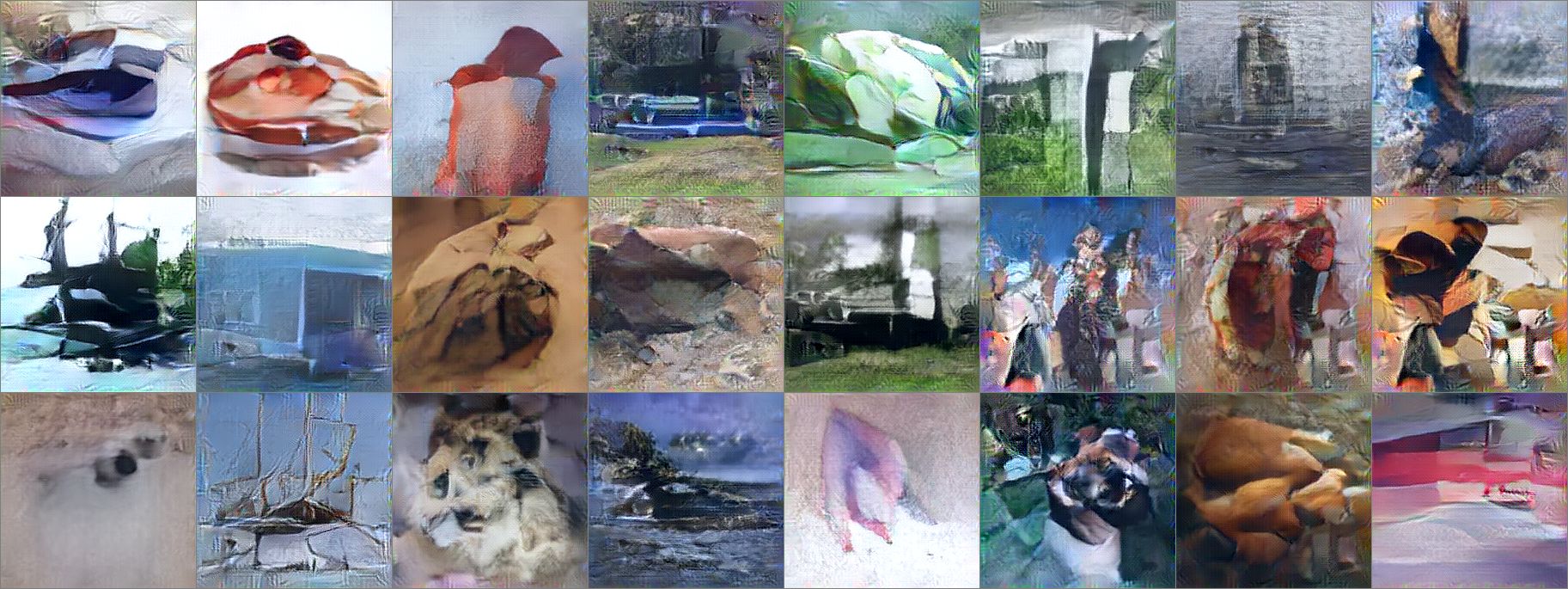}\\
\includegraphics[width=0.95\linewidth]{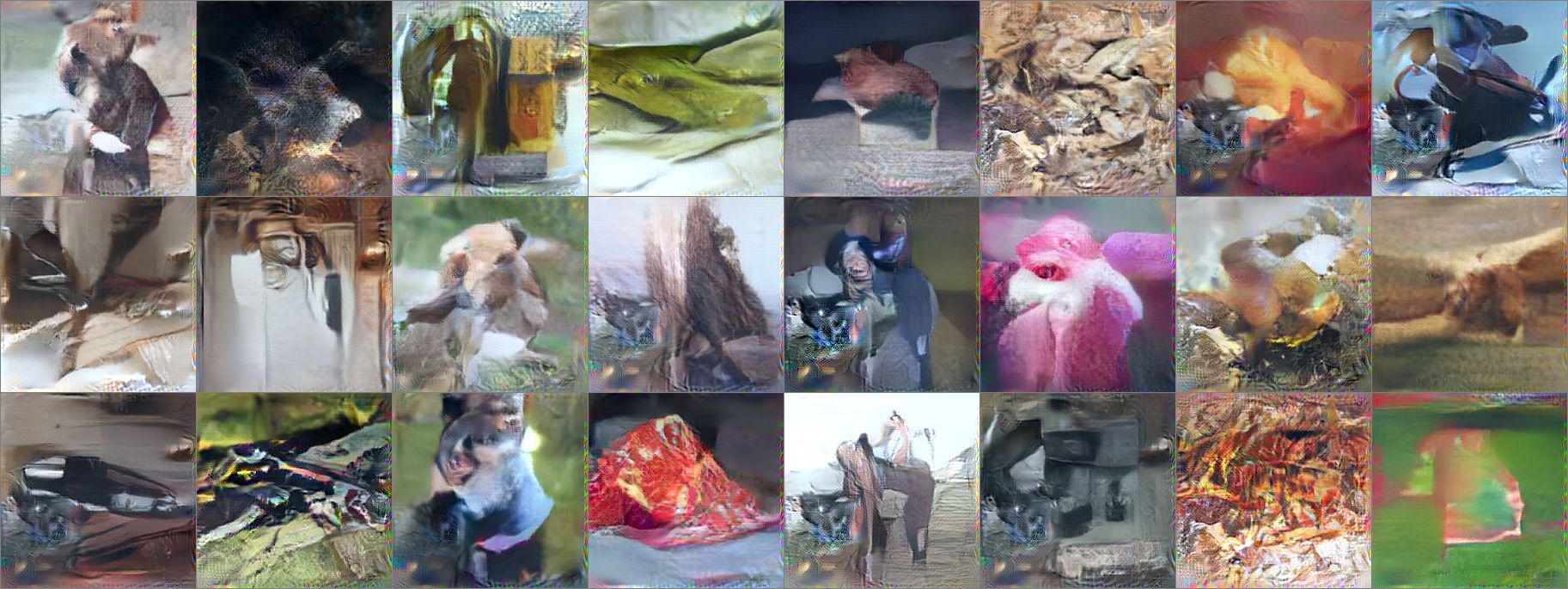}\\
\includegraphics[width=0.95\linewidth]{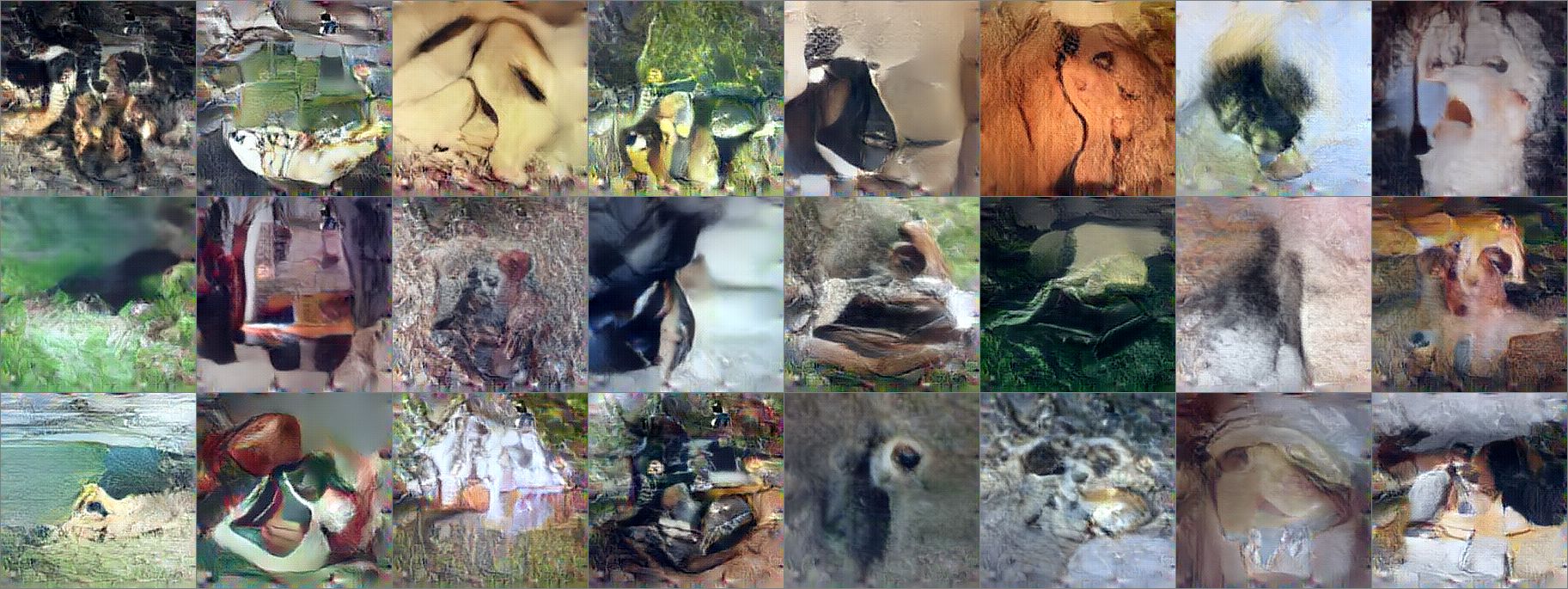}
\end{tabular}
\end{center}
\caption{Samples from VAE with the SE loss (\textbf{topmost}) and the proposed \ourapproach loss (\textbf{top to bottom:} AlexNet \conv5, AlexNet \fc6, VideoNet \conv5).}
\label{fig:vae_samples}
\end{figure}

\subsection{Variational autoencoder} \label{sec:exp_vae}
A standard VAE consists of an encoder $\enc$ and a decoder $\dec$.
The encoder maps an input sample $\vaein$ to a distribution over latent variables $\latent \sim \enc(\vaein) = q(\latent|\vaein)$.
$\dec$ maps from this latent space to a distribution over images $\vaerec \sim \dec(\latent) = p(\vaein|\latent)$.
The loss function is 
\begin{equation}\label{eq:vae}
\sum\limits_i -\expect_{q(\latent|\vaein_i)}\, \log p(\vaein_i|\latent) + \kldiv (q(\latent| \vaein_i) || p(\latent)),
\end{equation}
where $p(\latent)$ is a prior distribution of latent variables and $\kldiv$ is the Kullback-Leibler divergence.
The first term in Eq.~\ref{eq:vae} is a reconstruction error.
If we assume that the decoder predicts a Gaussian distribution at each pixel, then it reduces to squared Euclidean error in the image space.
The second term pulls the distribution of latent variables towards the prior.
Both $q(\latent|\vaein)$ and $p(\latent)$ are commonly assumed to be Gaussian, in which case the $KL$ divergence can be computed analytically.
Please refer to~\citet{Kingma_NIPS2014} for details.

We use the proposed loss instead of the first term in Eq.~\ref{eq:vae}.
This is similar to~\citet{Larsen_arxiv2015}, but the comparator does not have to be a part of the discriminator.
Technically, there is little difference from training an autoencoder.
First, instead of predicting a single latent vector $z$ we predict two vectors $\vaemean$ and $\vaesigma$ and sample $z = \vaemean + \vaesigma \odot \vaenoise$, where $\vaenoise$ is standard Gaussian (zero mean, unit variance) and $\odot$ is element-wise multiplication.
Second, we add the KL divergence term to the loss:
\begin{equation}
 \klloss = \frac{1}{2}\sum\limits_i \left(||\vaemean_i||_2^2   + ||\vaesigma_i||^2_2 - \langle \log \vaesigma_i^2, \, \textbf{1} \rangle \right) .
\end{equation}
We manually set the weighting of the KL term relative to the rest of the loss. 
Proper probabilistic derivation is non-straightforward, and we leave it for future research.

We trained on $227 \times 227$ pixel crops of $256 \times 256$ pixel ILSVRC-2012 images.
The encoder architecture is the same as AlexNet up to layer \fc6, and the decoder architecture is shown in Table~\ref{tbl:generator_arch}.
We initialized the encoder with AlexNet weights, however, this is not necessary, as shown in the appendix.
We sampled from the model by sampling the latent variables from a standard Gaussian $z = \vaenoise$ and generating images from that with the decoder.

Samples generated with the usual \ltwo loss, as well as three different comparators (AlexNet \conv5, AlexNet \fc6, VideoNet \conv5) are shown in Fig.~\ref{fig:vae_samples}.
While Euclidean loss leads to very blurry samples, our method yields images with realistic statistics.
Interestingly, the samples trained with the VideoNet comparator look qualitatively similar to the ones with AlexNet, showing that supervised training may not be necessary to yield a good comparator.
More results are shown in the appendix.

\begin{table}
   \begin{center}
   \setlength{\tabcolsep}{0.15cm}
  \small{
  \begin{tabular}{|l|cccccc|}
      \hline
      \small{Type}   & fc     & fc     & fc     & reshape & uconv         & conv  \\ \hline
      \small{InSize} & $-$    & $-$    & $-$    & $1$     & $4$            & $8$   \\
      \small{OutCh}  & $4096$ & $4096$ & $4096$ & $256$   & $256$          & $512$ \\
      \small{Kernel} & $-$    & $-$    & $-$    & $-$     & $4$            & $3$   \\
      \small{Stride} & $-$    & $-$    & $-$    & $-$     & $\uparrow \!2$ & $1$   \\
      \hline
    \end{tabular}}
\vspace{0.1cm}
   \setlength{\tabcolsep}{0.1cm}
  \small{
  \begin{tabular}{|l|ccccccc|}
      \hline
      \small{Type}   & uconv            & conv  & uconv          & conv  & uconv          & uconv            & uconv          \\ \hline
      \small{InSize} & $8$              & $16$  & $16$           & $32$  & $32$           & $64$             & $128$          \\
      \small{OutCh}  & $256$            & $256$ & $128$          & $128$ & $64$           & $32$             & $3$          \\
      \small{Kernel} & $4$              & $3$   & $4$            & $3$   & $4$            & $4$              & $4$            \\
      \small{Stride} & $\uparrow \!2$   & $1$   & $\uparrow \!2$ & $1$   & $\uparrow \!2$ & $\uparrow \!2$   & $\uparrow \!2$ \\
      \hline
    \end{tabular}}
  \end{center}
  \caption{Generator architecture for inverting layer \fc6 of AlexNet. }
  \label{tbl:generator_arch}
\end{table}

\begin{figure*}
\begin{center}
\setlength{\tabcolsep}{0.05cm}
\renewcommand{\arraystretch}{1}
  \begin{tabular}{cc}
  Image &
  \raisebox{-.5\height}{\includegraphics[width=0.92\linewidth]{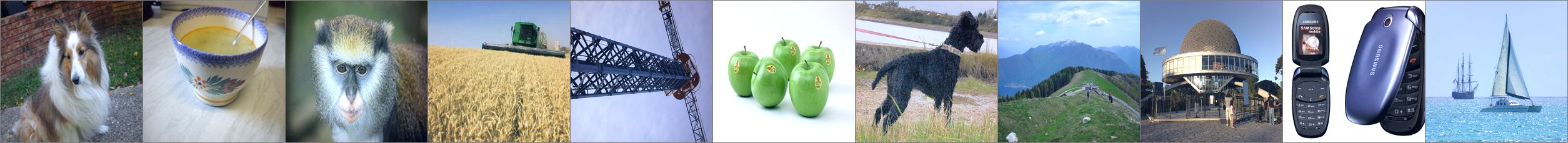}} \\
  \conv5 &
  \raisebox{-.5\height}{\includegraphics[width=0.92\linewidth]{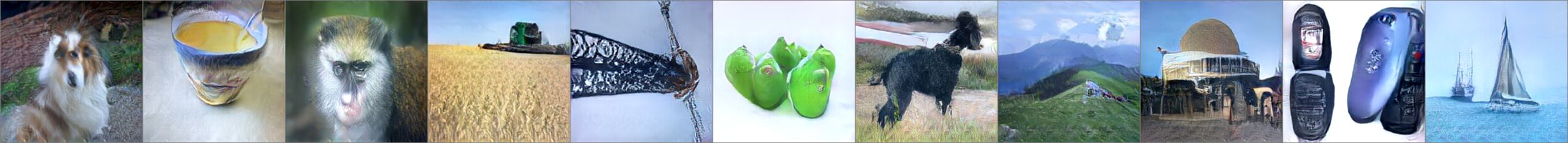}} \\ 
  \fc6 &
  \raisebox{-.5\height}{\includegraphics[width=0.92\linewidth]{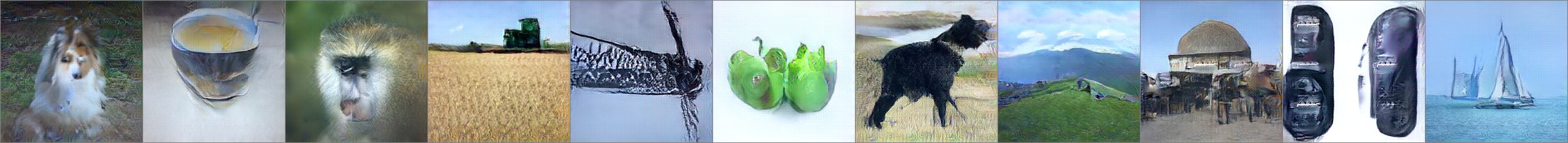}} \\ 
  \fc7 &
  \raisebox{-.5\height}{\includegraphics[width=0.92\linewidth]{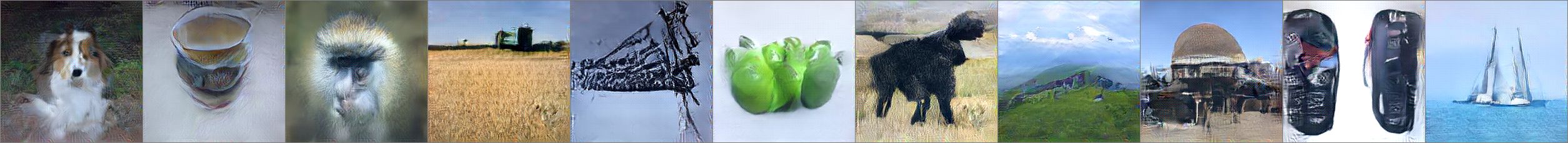}} \\ 
  \fc8 &
  \raisebox{-.5\height}{\includegraphics[width=0.92\linewidth]{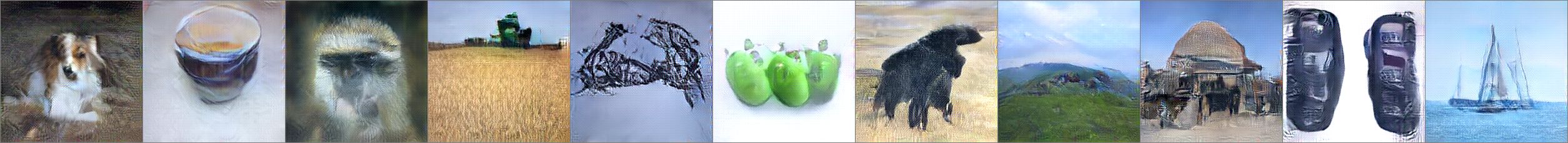}} \\ 
   \end{tabular}
\end{center}
   \caption{Representative reconstructions from higher layers of AlexNet. 
   General characteristics of images are preserved very well.
   In some cases (simple objects, landscapes) reconstructions are nearly perfect even from \fc8.
   In the leftmost column the network generates dog images from \fc7 and \fc8.
   }
\label{fig:AlexNet_recons}
\end{figure*}

\subsection{Inverting AlexNet} \label{sec:exp_inversion}
Analysis of learned representations is an important but largely unsolved problem.
One approach is to invert the representation.
This may give insights into which information is preserved in the representation and what are its invariance properties.
However, inverting a non-trivial feature representation $\repres$, such as the one learned by a large convolutional network, is a difficult ill-posed problem.

Our proposed approach inverts the AlexNet convolutional network very successfully.
Surprisingly rich information about the image is preserved in deep layers of the network and even in the predicted class probabilities.
While being an interesting result in itself, this also shows how \ourapproach is an excellent loss function when dealing with very difficult image restoration tasks.

Suppose we are given a feature representation $\repres$, which we aim to invert, and an image $\img$.
There are two inverse mappings: $\repres^{-1}_R$ such that $\repres(\repres^{-1}_R(\feat)) \approx \feat$, and $\repres^{-1}_L$ such that $\repres^{-1}_L (\repres (\img)) \approx \img$.
Recently two approaches to inversion have been proposed, which correspond to these two variants of the inverse.

\citet{Mahendran_CVPR2015}, as well as \citet{Simonyan_ICLR2014} and ~\citet{Yosinski_2015}, apply gradient-based optimization to find an image $\recimg$ which minimizes the loss 
\begin{equation}
 ||\repres(\img) - \repres(\recimg)||_2^2 + \prior (\recimg),
\end{equation}
where $\prior$ is a simple natural image prior, such as total variation (TV) regularizer.
This method produces images which are roughly natural and have features similar to the input features, corresponding to $\repres^{-1}_R$.
However, the prior is limited, so reconstructions from fully connected layers of AlexNet do not look much like natural images.

\citet{our_inverting} train up-convolutional networks on a large training set of natural images to perform the inversion task.
They use \ltwo distance in the image space as loss function, which leads to approximating $\repres^{-1}_L$.
The networks learn to reconstruct the color and rough positions of objects well, but produce over-smoothed results because they average all potential reconstructions.

Our method can be seen as combining the best of both worlds.
Loss in the feature space helps preserve perceptually important image features.
Adversarial training keeps reconstructions realistic.
Note that similar to \citet{our_inverting} and unlike \citet{Mahendran_CVPR2015}, our method 
does not require the feature representation being inverted to be differentiable.

\begin{figure}
\begin{center}
\setlength{\tabcolsep}{0.1cm}
\renewcommand{\arraystretch}{1}
  \begin{tabular}{cc}
  & Image \quad \conv5 \quad\; \fc6 \quad\;\;\, \fc7 \quad\;\;\; \fc8\;\; \\
  Our &
  \raisebox{-.5\height}{\includegraphics[width=0.85\linewidth]{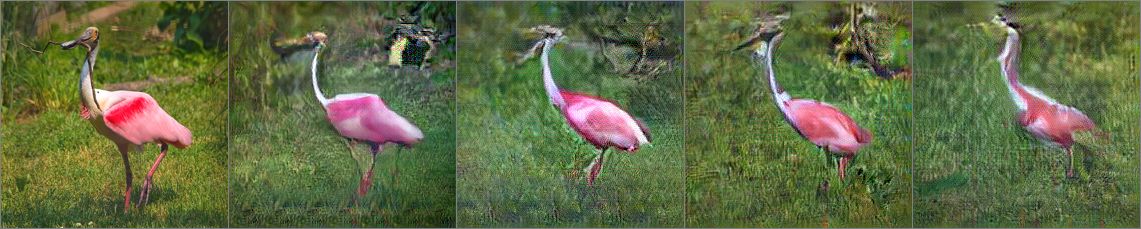}} \\ 
  D\&B &
  \raisebox{-.5\height}{\includegraphics[width=0.85\linewidth]{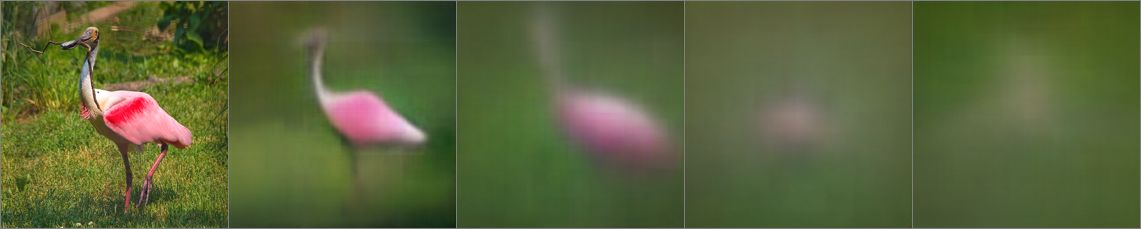}} \\
  M\&V &
  \raisebox{-.5\height}{\includegraphics[width=0.85\linewidth]{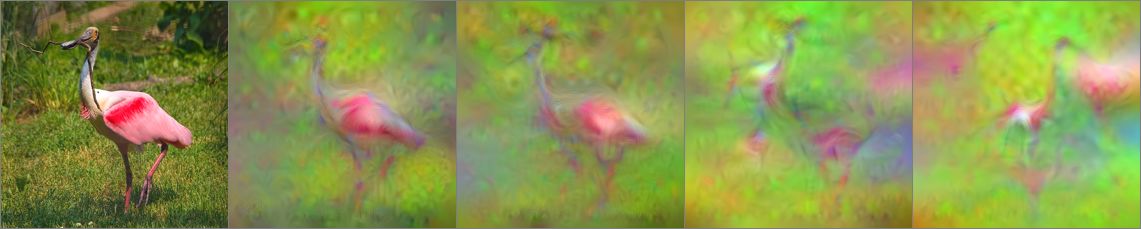}}\\
  \multicolumn{2}{c}{} \vspace*{-0.25cm} \\
   Our &
  \raisebox{-.5\height}{\includegraphics[width=0.85\linewidth]{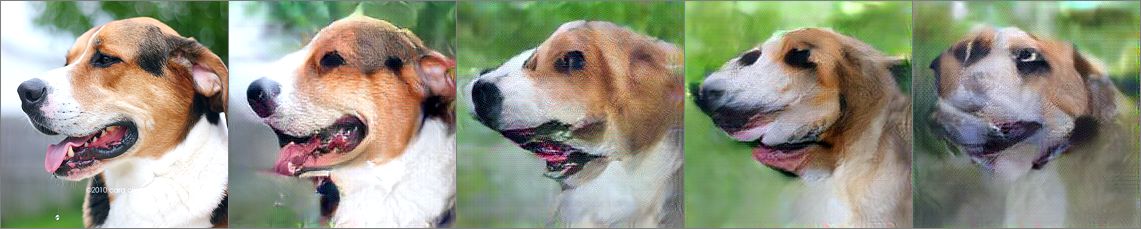}} \\ 
  D\&B &
  \raisebox{-.5\height}{\includegraphics[width=0.85\linewidth]{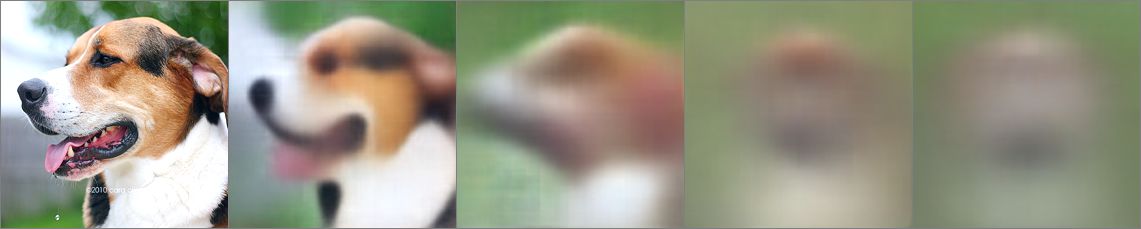}} \\
  M\&V &
  \raisebox{-.5\height}{\includegraphics[width=0.85\linewidth]{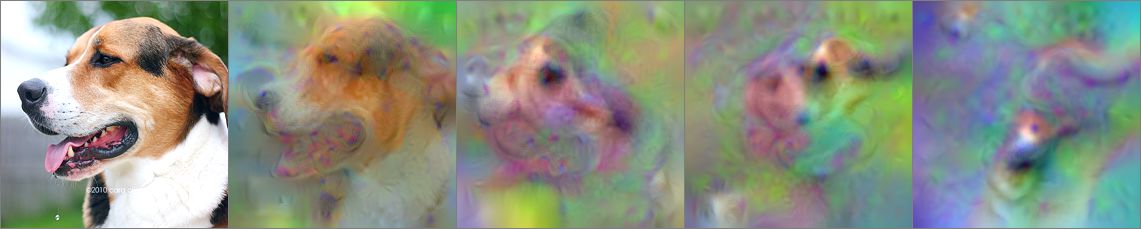}} \\
   \end{tabular}
\end{center}
   \caption{Comparison with \citet{our_inverting} and \citet{Mahendran_CVPR2015}.
   Our results look significantly better, even our failure cases (second image).}
\label{fig:AlexNet_comparison}
\end{figure}

\textbf{Technical details.}
The generator in this setup takes the features extracted by AlexNet and generates an image from them, that is, $\inp = \repres(\img),\, \targ = \img$.
In general we followed~\citet{our_inverting} in designing the generators.
The only modification is that we inserted more convolutional layers, giving the network more capacity.
We reconstruct from outputs of layers \conv5~--\fc8.
In each layer we also include processing steps following the layer, that is, pooling and non-linearities. 
So for example \conv5 means pooled features (pool5), and \fc6 means rectified values (relu6).

Architecture used for inverting \fc6 is the same as the decoder of the VAE shown in Table~\ref{tbl:generator_arch}.
Architectures for other layers are similar, except that for reconstruction from \conv5 fully connected layers are replaced by convolutional ones. 
The discriminator is the same as used for VAE.
We trained on the ILSVRC-2012 training set and evaluated on the ILSVRC-2012 validation set.

\begin{figure}
\begin{center}
\setlength{\tabcolsep}{0.1cm}
\renewcommand{\arraystretch}{1}
\small{
  \begin{tabular}{c}
  \;\;Image \quad\;\; Full \quad\; $-\pixloss$ \;\;\; $-\featloss$ \; $-\discrloss$  $\begin{array} {lcl} -\featloss \\ -\discrloss \end{array}$  \\
  \includegraphics[width=0.95\linewidth]{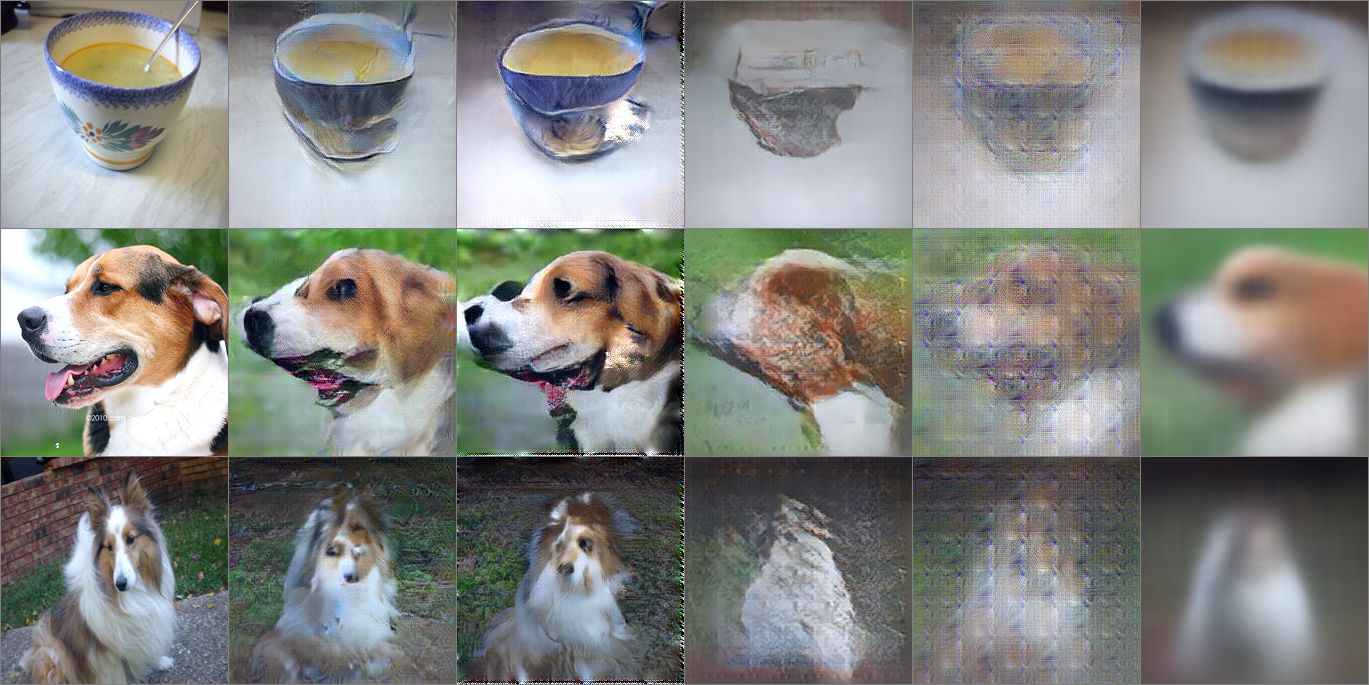}\\
   \end{tabular}}
\end{center}
   \caption{Reconstructions from \fc6 with some components of the loss removed.}
\label{fig:ablation}
\end{figure}

\textbf{Ablation study.}
We tested if all components of our loss are necessary.
Results with some of these components removed are shown in Fig.~\ref{fig:ablation}.
Clearly the full model performs best. In the following we will give some intuition why.

Training just with loss in the image space leads to averaging all potential reconstructions, resulting in over-smoothed images.
One might imagine that adversarial training would allow to make images sharp.
This indeed happens, but the resulting reconstructions do not correspond to actual objects originally contained in the image.
The reason is that any ``natural-looking'' image which roughly fits the blurry prediction minimizes this loss.
Without the adversarial loss predictions look very noisy.
Without the image space loss the method works well, but one can notice artifact on the borders of images, and training was less stable in this case.

\textbf{Sampling pre-images.}
Given a feature vector $\feat$, it would be interesting to sample multiple images $\recimg$ such that $\repres(\recimg) = \feat$.
A straightforward approach would inject noise into the generator along with the features, so that the network could randomize its outputs.
This does not yield the desired result, since nothing in the loss function forces the generator to output multiple different reconstructions per feature vector.
A major problem is that in the training data we only have one image per feature vector, i.e., a single sample per conditioning vector.
We did not attack this problem in our paper, but we believe it is an important research direction.

\textbf{Best results.}
Representative reconstructions from higher layers of AlexNet are shown in Fig.~\ref{fig:AlexNet_recons}.
Comparison with existing approaches is shown in Fig.~\ref{fig:AlexNet_comparison}.
Reconstructions from \conv5 are near-perfect, combining the natural colors and sharpness of details.
Reconstructions from fully connected layers are still very good, preserving the main features of images, colors, and positions of large objects.

Normalized Euclidean error in image space and in feature space (that is, the distance between the features of the image and the reconstruction) are shown in Table~\ref{tbl:inversion_quant}. 
The method of Mahendran\&Vedaldi performs well in feature space, but not in image space,
the method of Dosovitskiy\&Brox~--- vice versa.
The presented approach is fairly good on both metrics.

\begin{figure}
\begin{center}
\setlength{\tabcolsep}{0.15cm}
\renewcommand{\arraystretch}{1}
\small{
  \begin{tabular}{l|cccc|}
                              & \conv5   & \fc6     & \fc7      & \fc8        \\ \hline
  Mahendran\&Vedaldi          & $71/19$  & $80/19$  & $82/16$   & $84/09$    \\  
  Dosovitskiy \& Brox         & $35/ - $ & $51/ - $ & $56/ - $  & $58/ -  $    \\ \hline
  Our just image loss         & $ -/ -$  & $46/79$  & $ - / - $ & $ - / - $    \\
  Our AlexNet \conv5          & $43/37$  & $55/48$  & $61/45$   & $63/29$    \\  
  Our VideoNet \conv5         & $ -/ -$  & $51/57$  & $ - / - $ & $ - / - $    \\   \hline
   \end{tabular}}
   \vspace{-0.2cm}
\end{center}
   \caption{Normalized inversion error (in \%) when reconstructing from different layers of AlexNet with different methods.
            First in each pair~-- error in the image space, second~-- in the feature space.}
\label{tbl:inversion_quant}
\vspace{-0.2cm}
\end{figure}

\begin{figure}
\begin{center}
\setlength{\tabcolsep}{0.03cm}
\renewcommand{\arraystretch}{1}
  \begin{tabular}{ccccc}
  & \conv5 & \fc6 & \fc7 & \fc8 \\
  \renewcommand{\arraystretch}{2.1}
  \begin{tabular}{@{}c@{}}1\; \\ 2\; \\ 4\;\\ 8\; \end{tabular} &
  \renewcommand{\arraystretch}{1}
  \raisebox{-.5\height}{\includegraphics[width=0.22\linewidth]{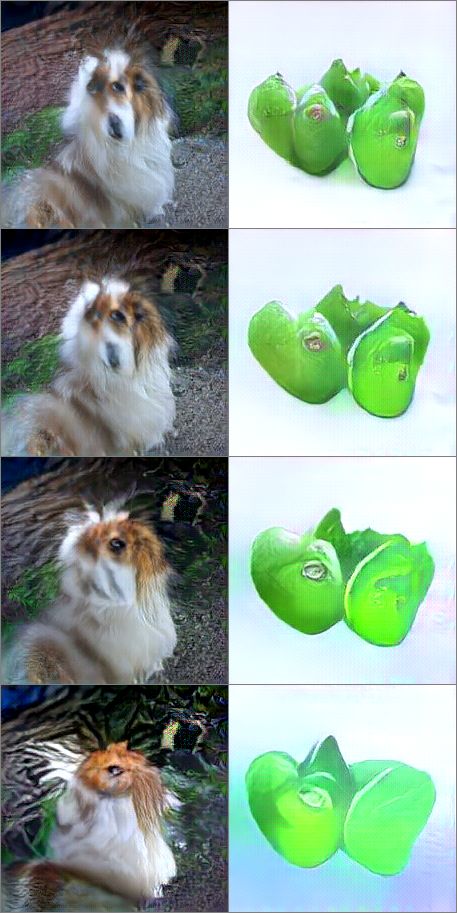}} &
  \raisebox{-.5\height}{\includegraphics[width=0.22\linewidth]{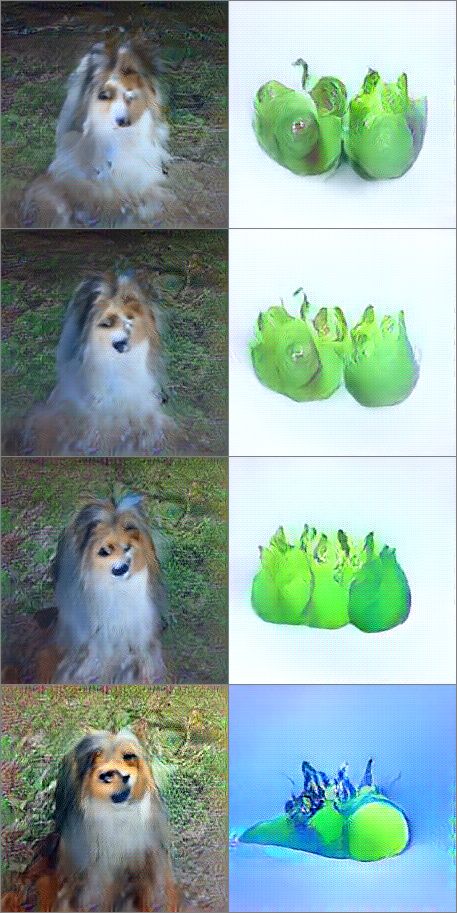}} &
  \raisebox{-.5\height}{\includegraphics[width=0.22\linewidth]{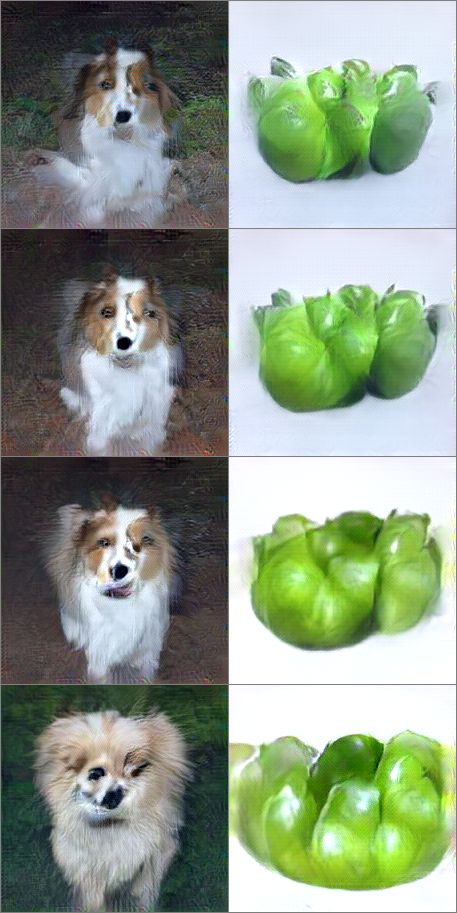}} &
  \raisebox{-.5\height}{\includegraphics[width=0.22\linewidth]{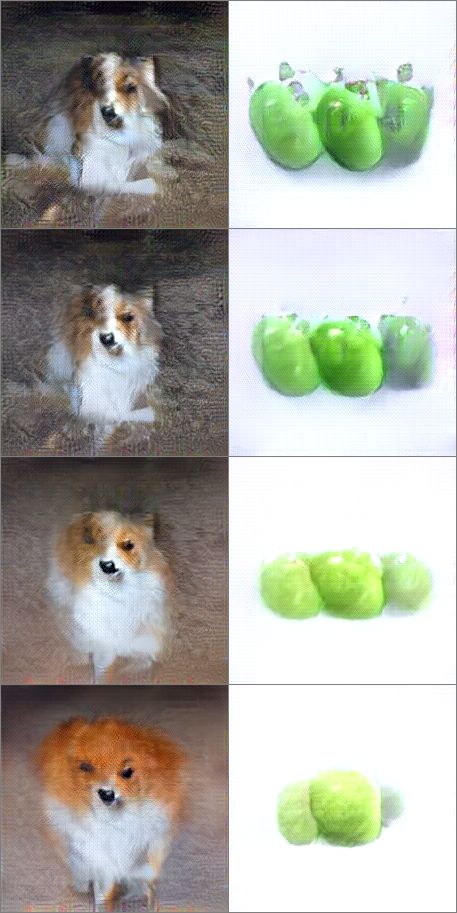}} 
   \end{tabular}
\end{center}
   \caption{Iteratively re-encoding images with AlexNet and reconstructing. Iteration number shown on the left.}
\label{fig:AlexNet_iterative}
\end{figure}

\textbf{Iterative re-encoding.}
We performed another experiment illustrating how similar are the features of reconstructions to the original image features.
Given an image, we compute its features, generate an image from those, and then iteratively compute the features of the result and generate from those.
Results are shown in Fig.~\ref{fig:AlexNet_iterative}.
Interestingly, several iterations do not significantly change the reconstruction, indicating that important perceptual features are preserved in the generated images. 
More results are shown in the appendix.

\begin{figure}
\begin{center}
\setlength{\tabcolsep}{0.1cm}
\renewcommand{\arraystretch}{1}
  \begin{tabular}{c}
  Image pair 1 \qquad\qquad\qquad\qquad Image pair 2 \\
  {\includegraphics[width=0.95\linewidth]{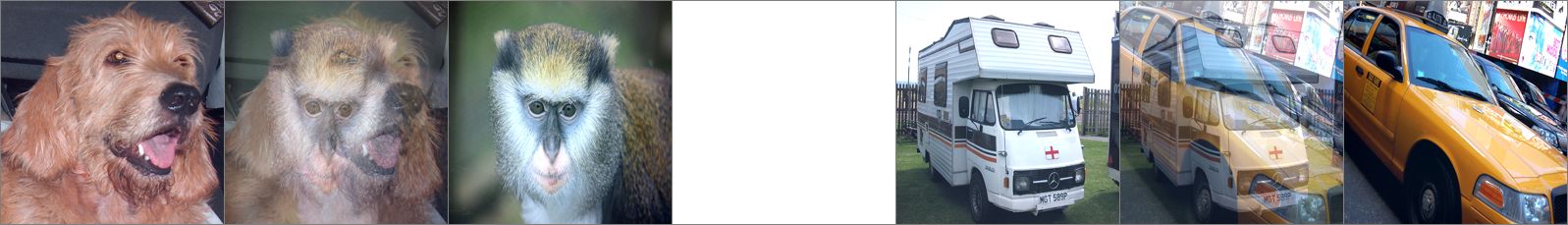}} \\
  \fc6 \\
  {\includegraphics[width=0.95\linewidth]{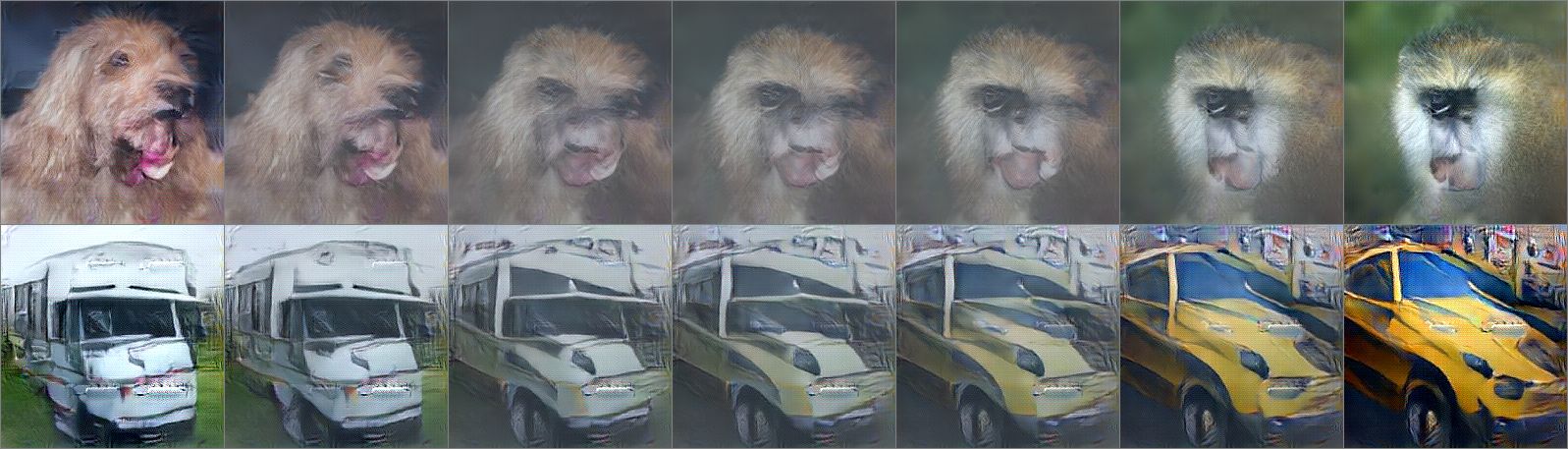}} \\ 
  \fc8 \\
  {\includegraphics[width=0.95\linewidth]{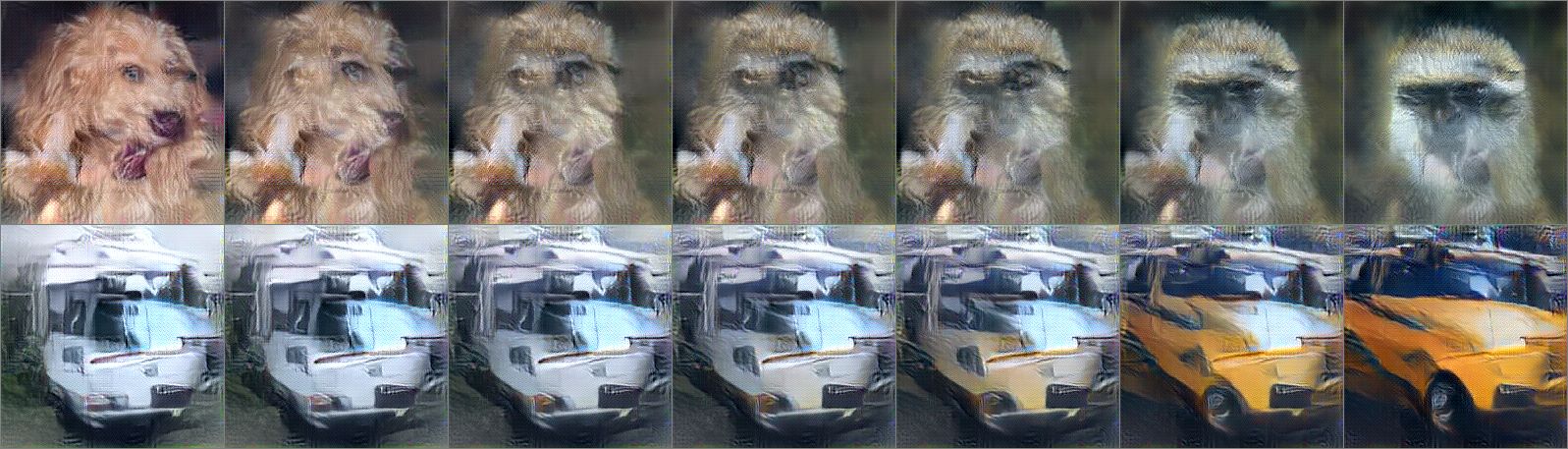}}
   \end{tabular}
\end{center}
   \caption{Interpolation between images by interpolating between their features in \fc6 and \fc8.}
\label{fig:AlexNet_interpol}
\end{figure}

\textbf{Interpolation.}
We can morph images into each other by linearly interpolating between their features and generating the corresponding images.
Fig.~\ref{fig:AlexNet_interpol} shows that objects shown in the images smoothly warp into each other.
More examples are shown in the appendix.

\textbf{Different comparators.}
AlexNet network we used above as comparator has been trained on a huge labeled dataset.
Is this supervision really necessary to learn a good comparator?
We show here results with several alternatives to \conv5 features of AlexNet: 1) $\fc6$ features of AlexNet, 2) \conv5 of AlexNet with random weights, 3) \conv5 of the network of~\citet{Wang_ICCV2015} which we refer to as VideoNet.

\begin{figure}
\begin{center}
\setlength{\tabcolsep}{0.1cm}
\renewcommand{\arraystretch}{1}
  \begin{tabular}{c}
  Image \;\;\;\; Alex5 \;\;\; Alex6 \;\;\; Video5 \;\;\;  Rand5 \\
  \includegraphics[width=0.85\linewidth]{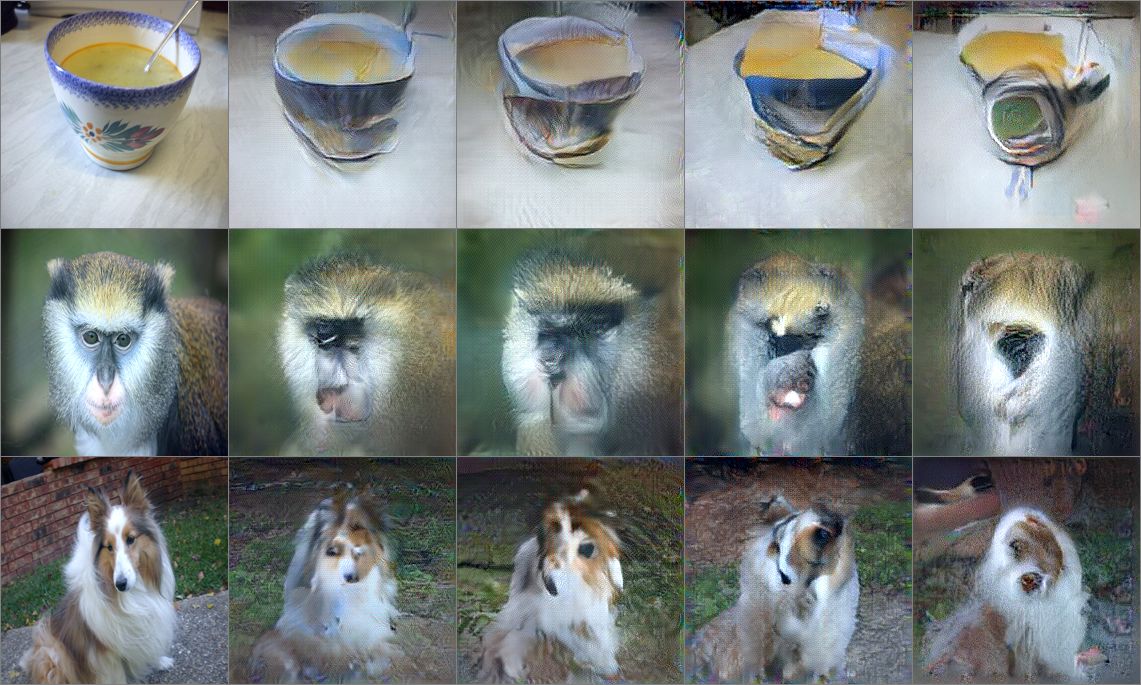}\\
   \end{tabular}
\end{center}
   \caption{Reconstructions from \fc6 with different comparators. The number indicates the layer from which features were taken.}
\label{fig:different_comparators}
\end{figure}

The results are shown in Fig.~\ref{fig:different_comparators}.
While AlexNet \conv5 comparator provides best reconstructions, other networks preserve key image features as well.
We also ran preliminary experiments with \conv5 features from the discriminator serving as a comparator, but were not able to get satisfactory results with those.

\section{Conclusion}

We proposed a class of loss functions applicable to image generation that are based on distances in feature spaces.
Applying these to three tasks~--- image auto-encoding, random natural image generation with a VAE and feature inversion~--- reveals that our loss is clearly superior to the typical loss in image space.
In particular, it allows reconstruction of perceptually important details even from very low-dimensional image representations.
We evaluated several feature spaces to measure distances.
More research is necessary to find optimal features to be used depending on the task.
To control the degree of realism in generated images, an alternative to adversarial training is an approach making use of feature statistics, similar to~\citet{Gatys_arxiv2015}. 
We see these as interesting directions of future work.

\section*{Acknowledgements}
The authors are grateful to Jost Tobias Springenberg and Philipp Fischer for useful discussions.
We acknowledge funding by the ERC Starting Grant VideoLearn (279401).

\bibliography{../dosovits_new}
\bibliographystyle{icml2016}


\section*{Appendix}

Here we show some additional results obtained with the proposed method.

\begin{figure}[h]
\begin{center}
\setlength{\tabcolsep}{0.1cm}
\renewcommand{\arraystretch}{1}
  \begin{tabular}{ccc}
  Image &
  \raisebox{-.5\height}{\includegraphics[width=0.4\linewidth]{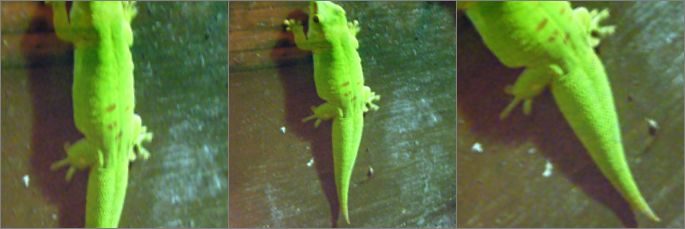}} &
  \raisebox{-.5\height}{\includegraphics[width=0.4\linewidth]{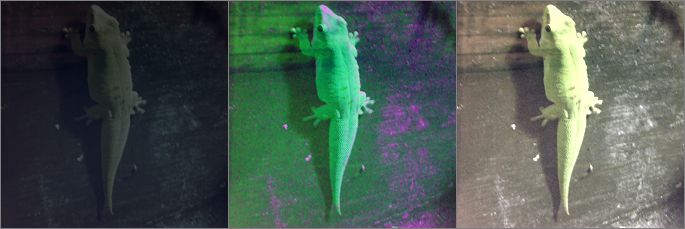}}\\
  \conv5 &
  \raisebox{-.5\height}{\includegraphics[width=0.4\linewidth]{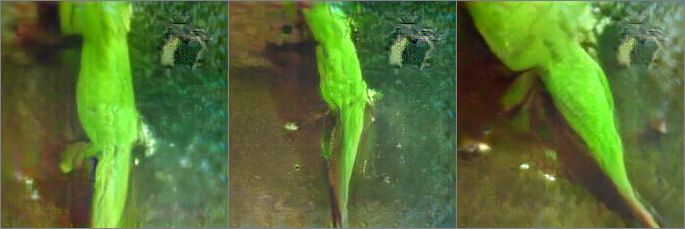}} &
  \raisebox{-.5\height}{\includegraphics[width=0.4\linewidth]{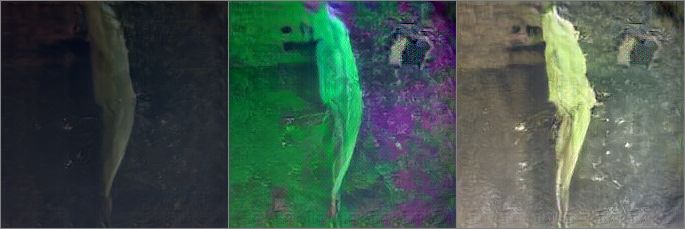}}\\ 
  \fc6 &
  \raisebox{-.5\height}{\includegraphics[width=0.4\linewidth]{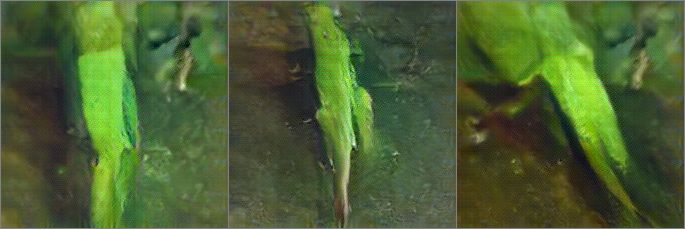}} &
  \raisebox{-.5\height}{\includegraphics[width=0.4\linewidth]{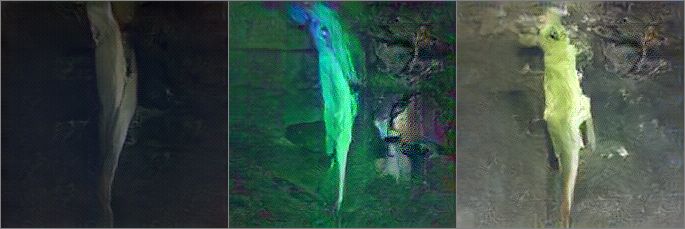}}\\ 
  \fc7 &
  \raisebox{-.5\height}{\includegraphics[width=0.4\linewidth]{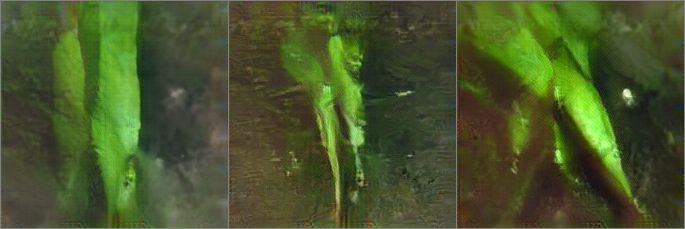}} &
  \raisebox{-.5\height}{\includegraphics[width=0.4\linewidth]{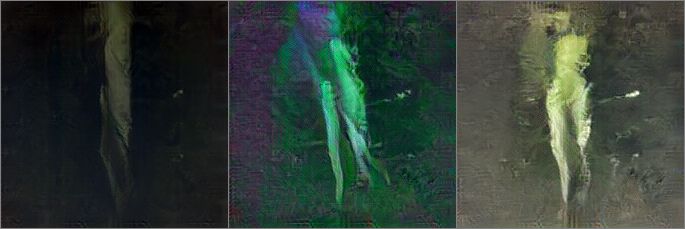}}\\ 
  \fc8 &
  \raisebox{-.5\height}{\includegraphics[width=0.4\linewidth]{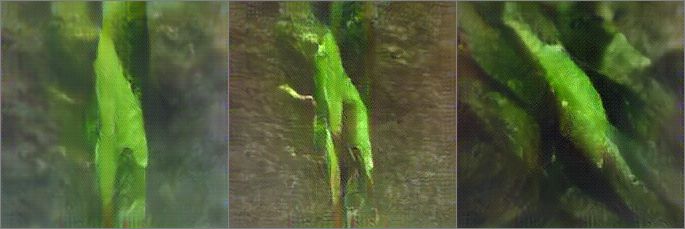}} &
  \raisebox{-.5\height}{\includegraphics[width=0.4\linewidth]{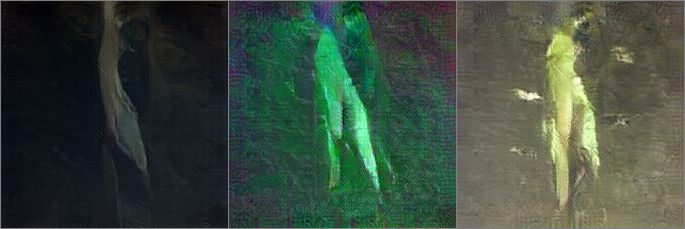}}\\ 
   \end{tabular}
\end{center}
   \caption{Position (first three columns) and color (last three columns) preservation.}
\label{fig:AlexNet_position}
\end{figure}

Figure~\ref{fig:AlexNet_position} illustrates how position and color of an object is preserved in deep layers of AlexNet~\citep{Krizhevsky_NIPS2012}.

\begin{figure*}
\begin{center}
\setlength{\tabcolsep}{0.1cm}
\begin{tabular}{cc}
\includegraphics[width=0.48\linewidth]{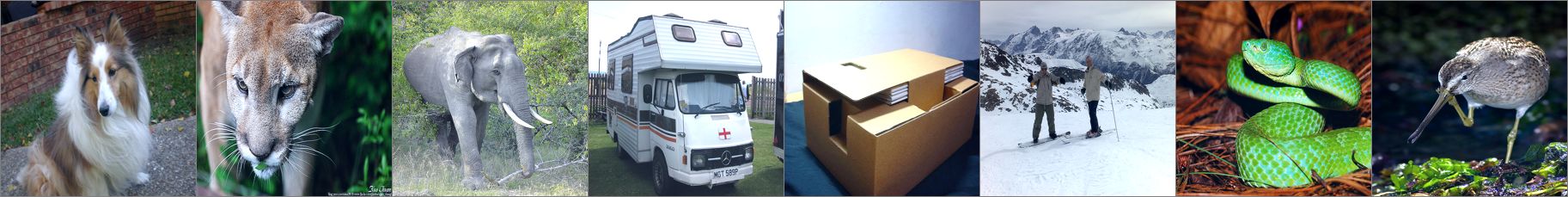} &
\includegraphics[width=0.48\linewidth]{ICML_orig_interpolate_supp.jpg} \\
\includegraphics[width=0.48\linewidth]{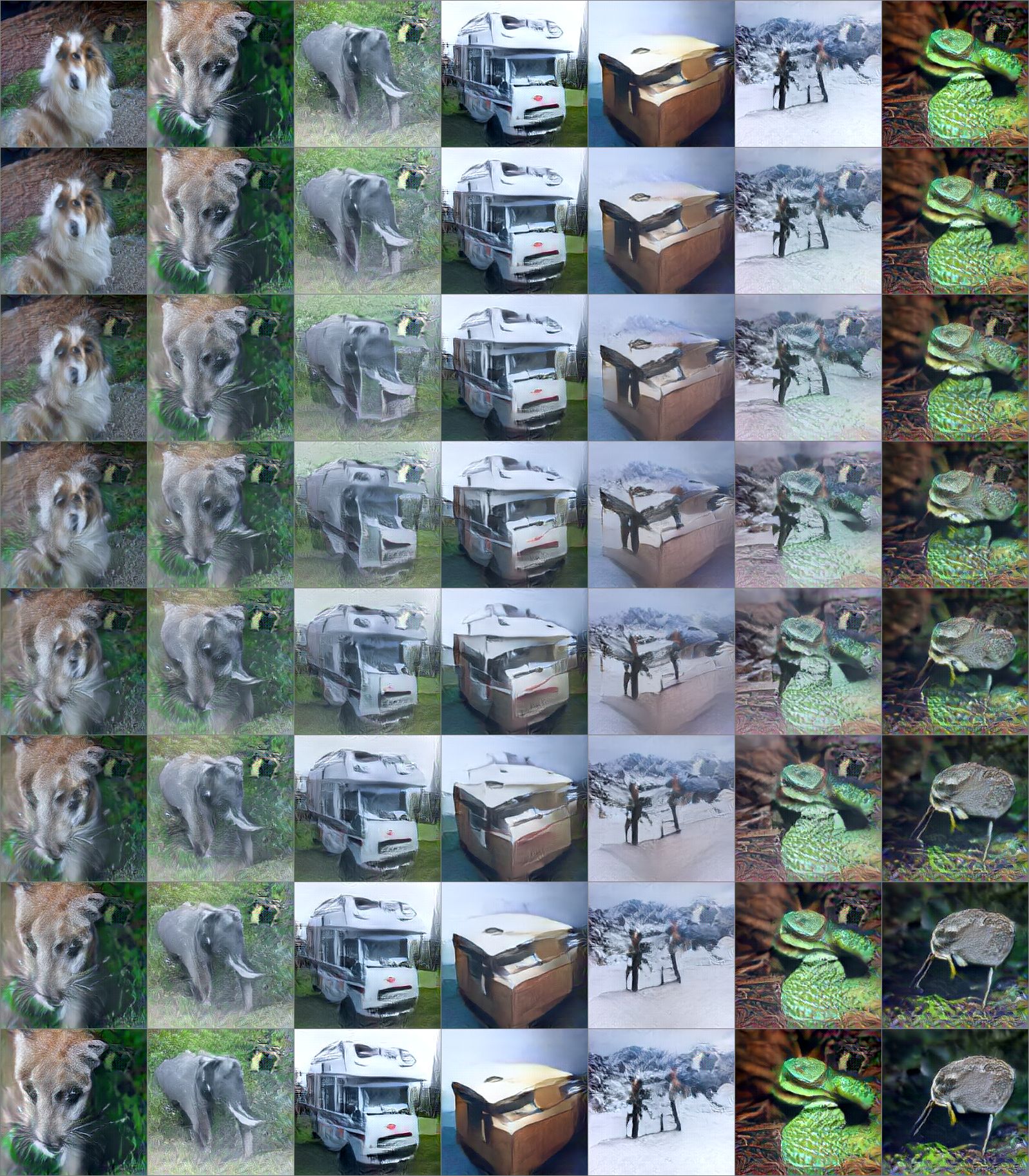} &
\includegraphics[width=0.48\linewidth]{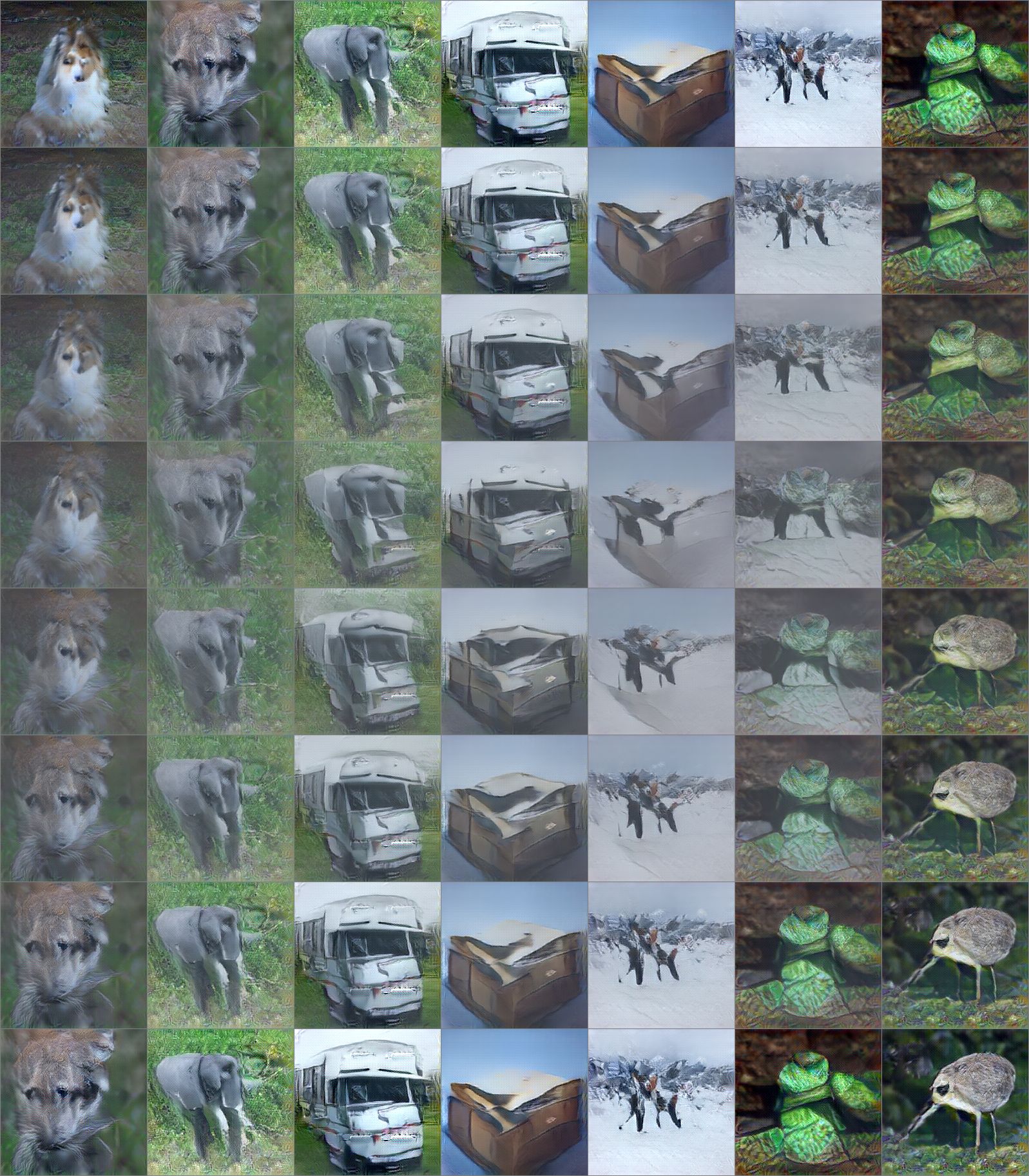} \\
\includegraphics[width=0.48\linewidth]{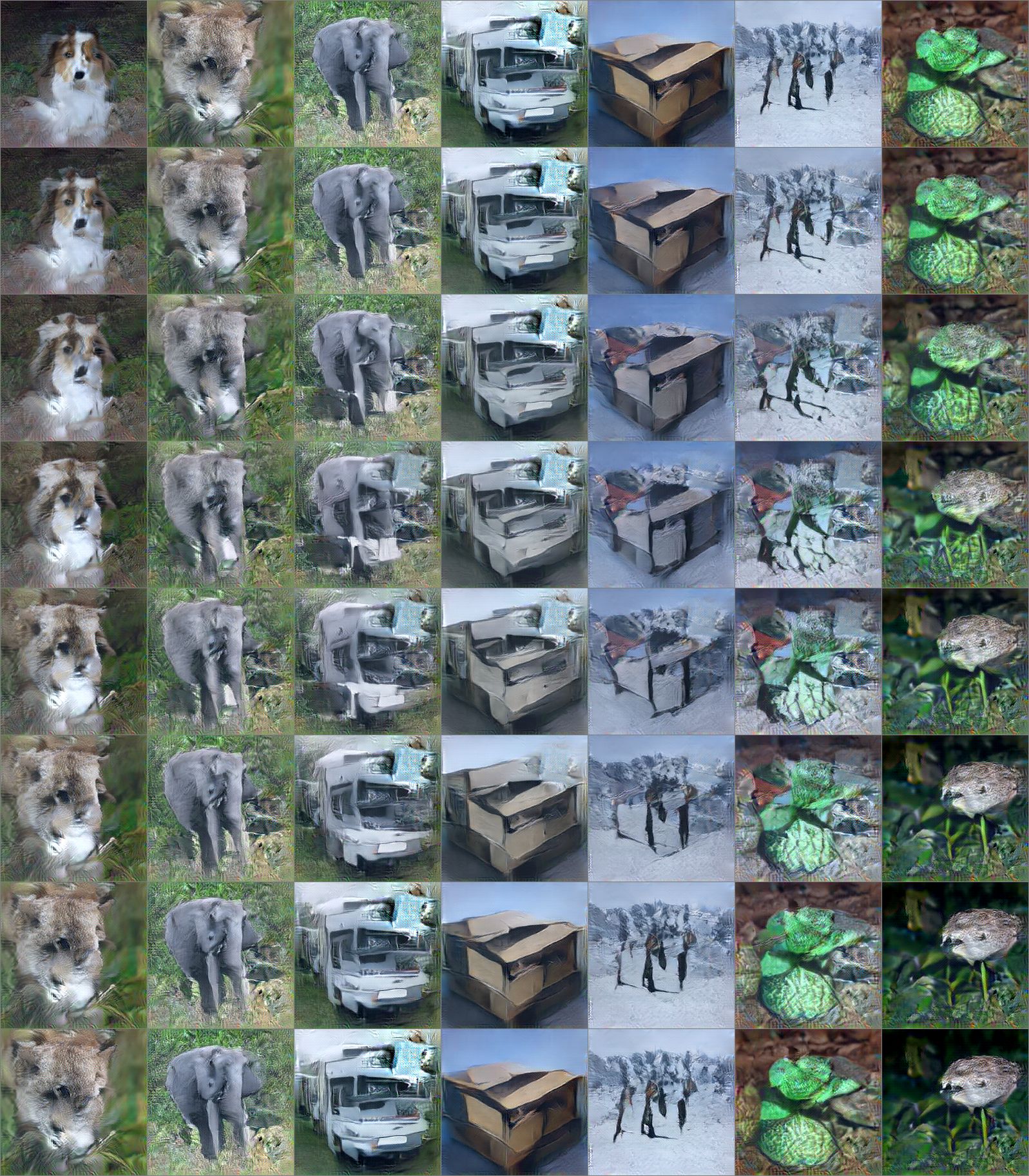} &
\includegraphics[width=0.48\linewidth]{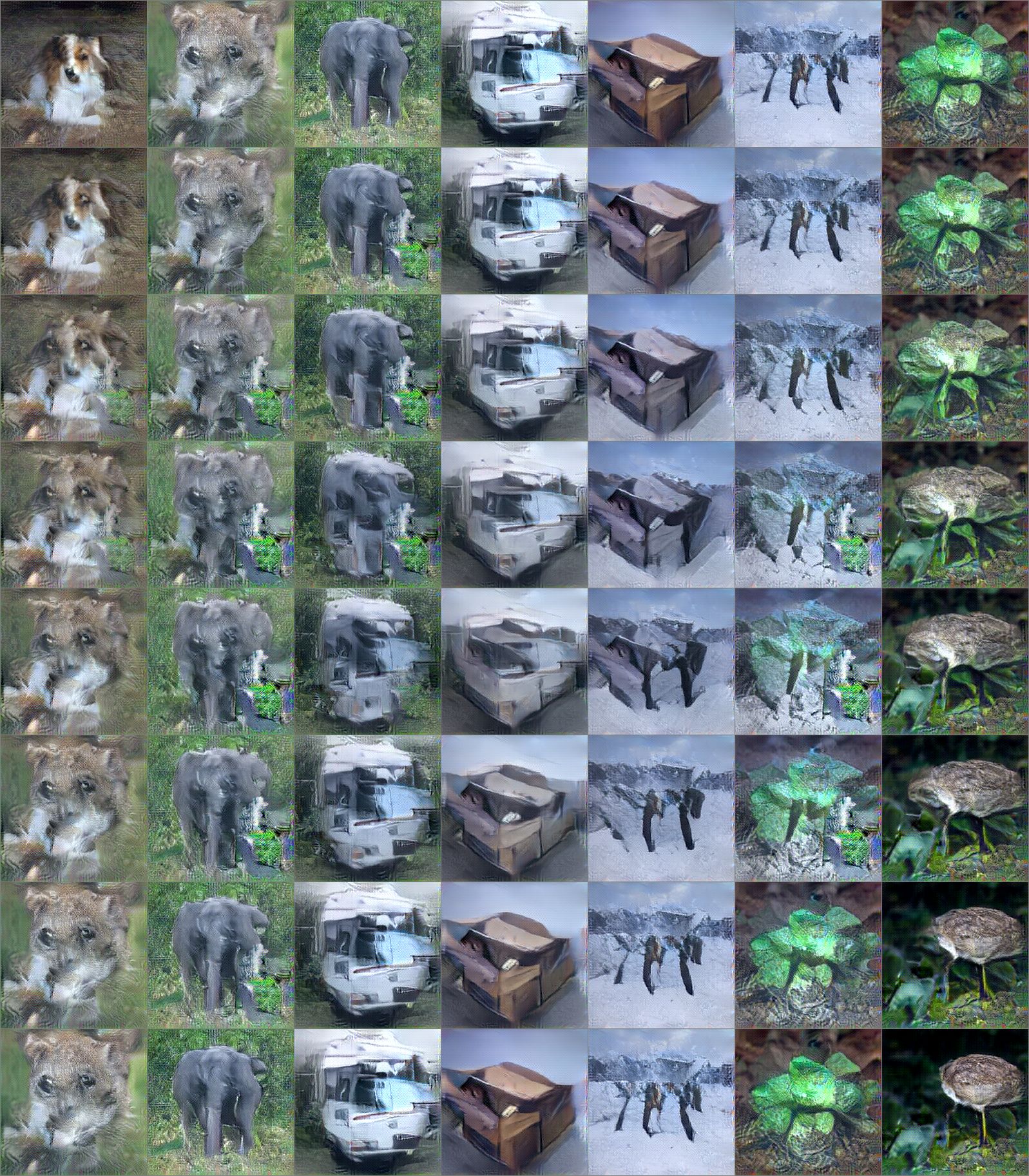}
\end{tabular}
\end{center}
\caption{Interpolation in feature spaces at different layers of AlexNet.
\textbf{Topmost:} input images,
\textbf{Top left:} \conv5,
\textbf{Top right:} \fc6,
\textbf{Bottom left:} \fc7,
\textbf{Bottom right:} \fc8.}
\label{fig:interpol}
\end{figure*}

Figure~\ref{fig:interpol} shows results of generating images from interpolations between the features of natural images.

\begin{figure*}
\begin{center}
\setlength{\tabcolsep}{0.1cm}
\begin{tabular}{cc}
\includegraphics[width=0.48\linewidth]{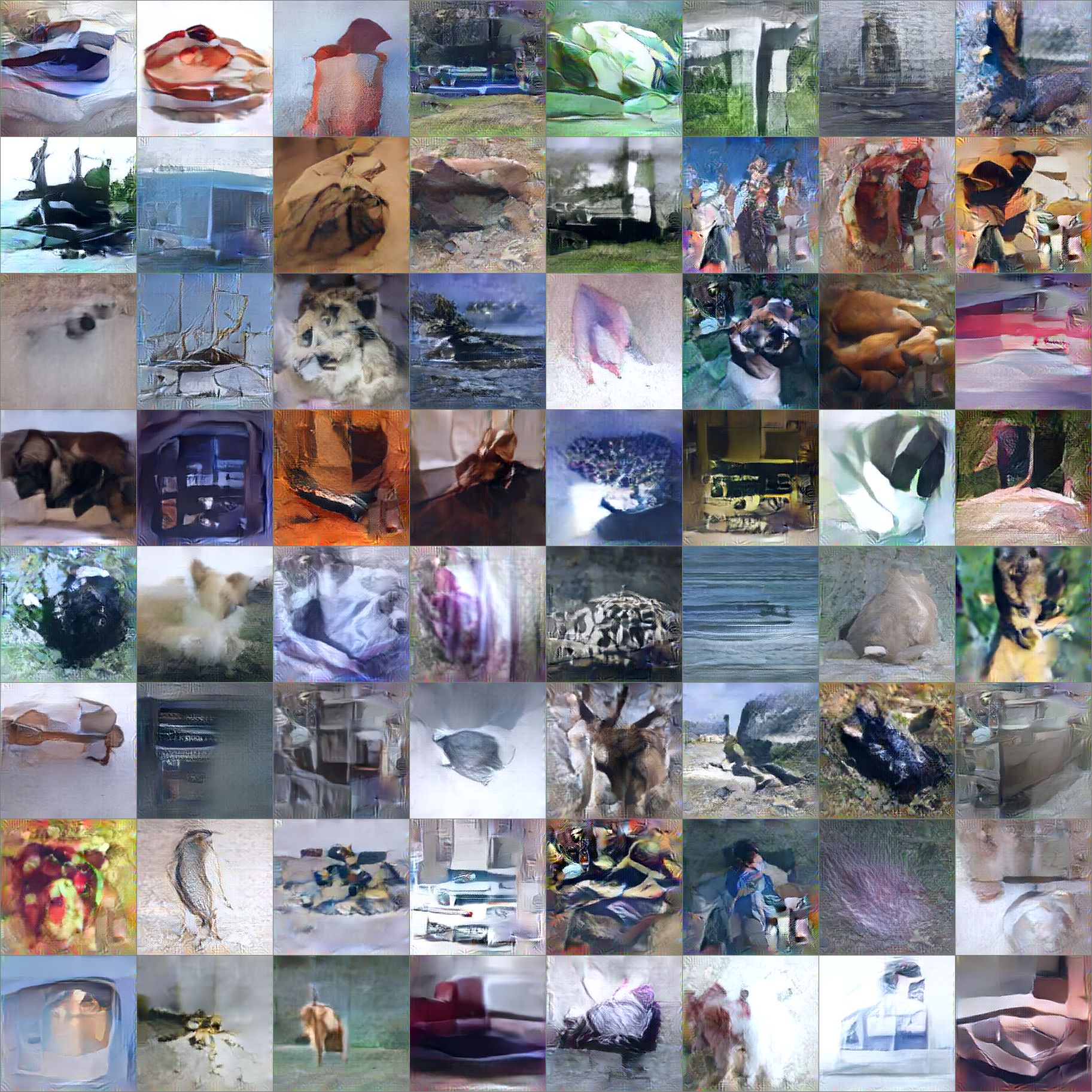} &
\includegraphics[width=0.48\linewidth]{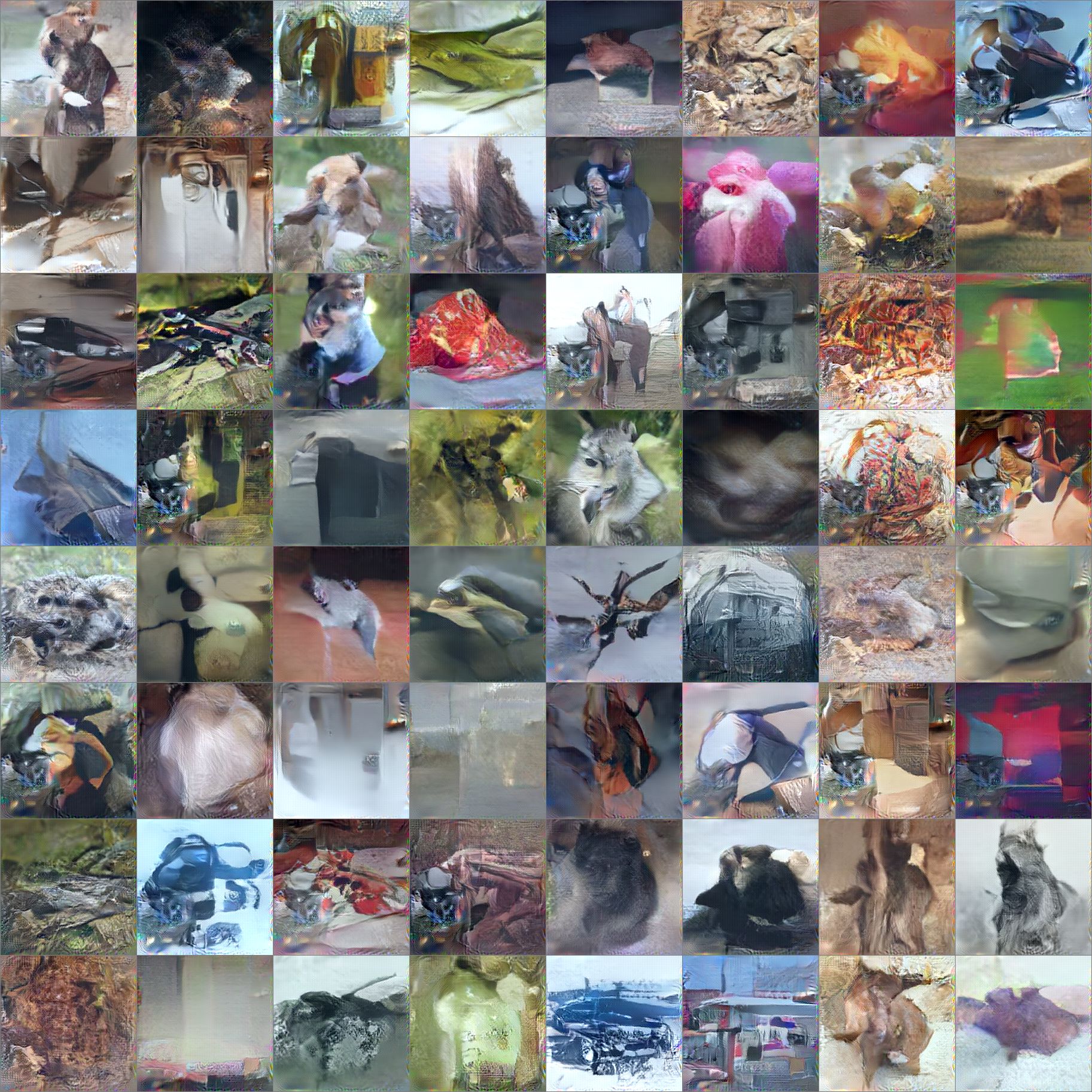} \\
\includegraphics[width=0.48\linewidth]{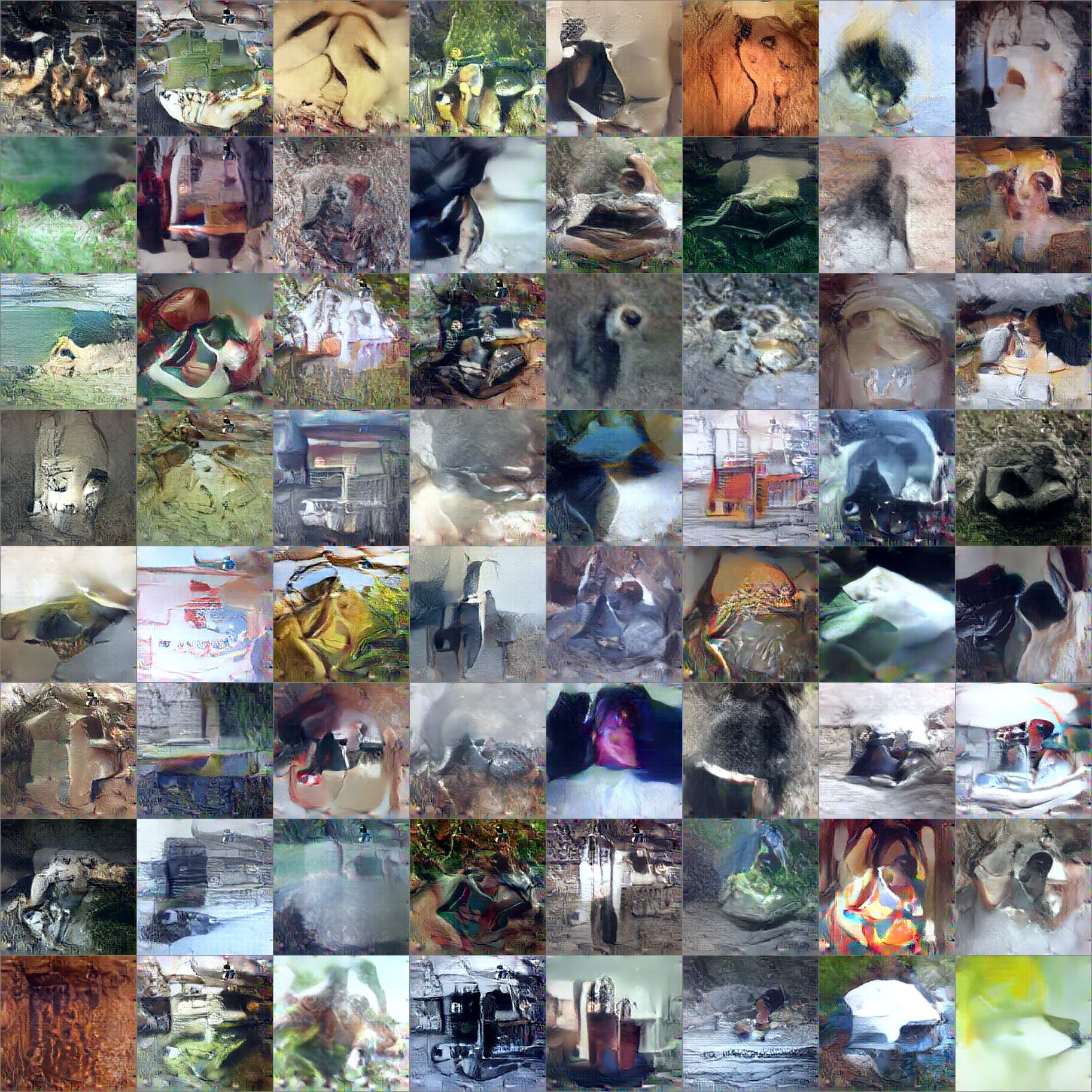} &
\includegraphics[width=0.48\linewidth]{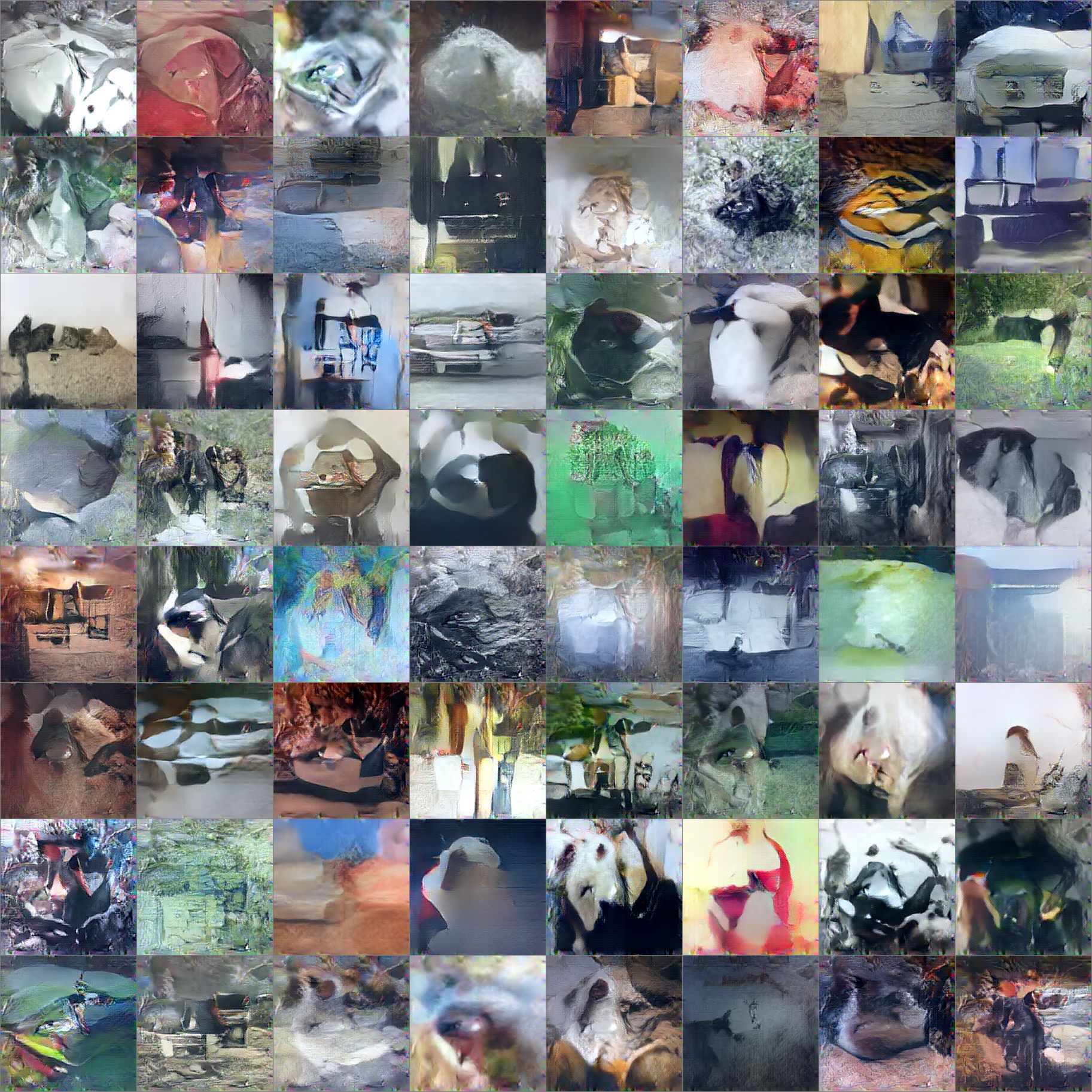}
\end{tabular}
\end{center}
\caption{Samples from VAE with our approach, with different comparators.
\textbf{Top left:} AlexNet \conv5 comparator,
\textbf{Top right:} AlexNet \fc6 comparator,
\textbf{Bottom left:} VideoNet \conv5 comparator,
\textbf{Bottom right:} VideoNet \conv5 comparator with randomly initialized encoder.}
\label{fig:vae_samples_supp}
\end{figure*}

Figure~\ref{fig:vae_samples_supp} shows samples from variational autoencoders with different losses.
Fully unsupervised VAE with VideoNet~\citep{Wang_ICCV2015} loss and random initialization of the encoder is in the bottom right.
Samples from this model are qualitatively similar to others, showing that initialization with AlexNet is not necessary.

\begin{figure*}
\begin{center}
\setlength{\tabcolsep}{0.1cm}
\begin{tabular}{cc}
\includegraphics[width=0.48\linewidth]{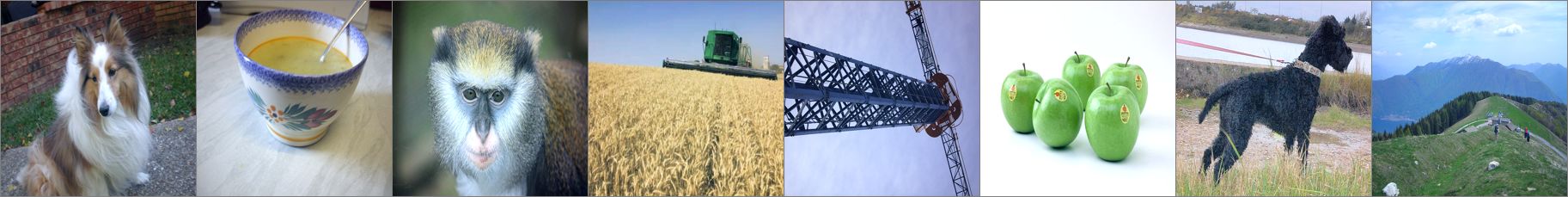} &
\includegraphics[width=0.48\linewidth]{ICML_orig_iterative_supp.jpg} \\
\includegraphics[width=0.48\linewidth]{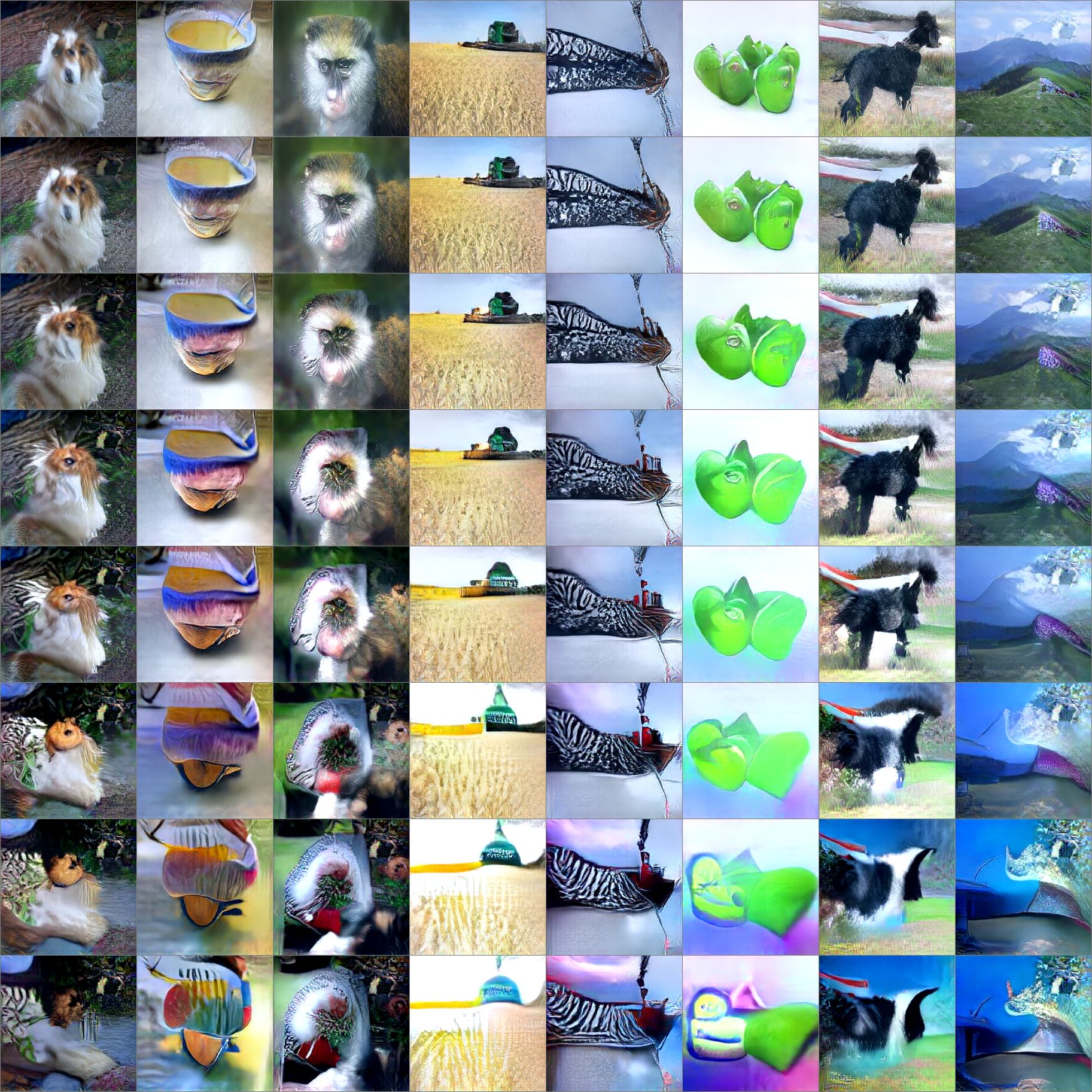} &
\includegraphics[width=0.48\linewidth]{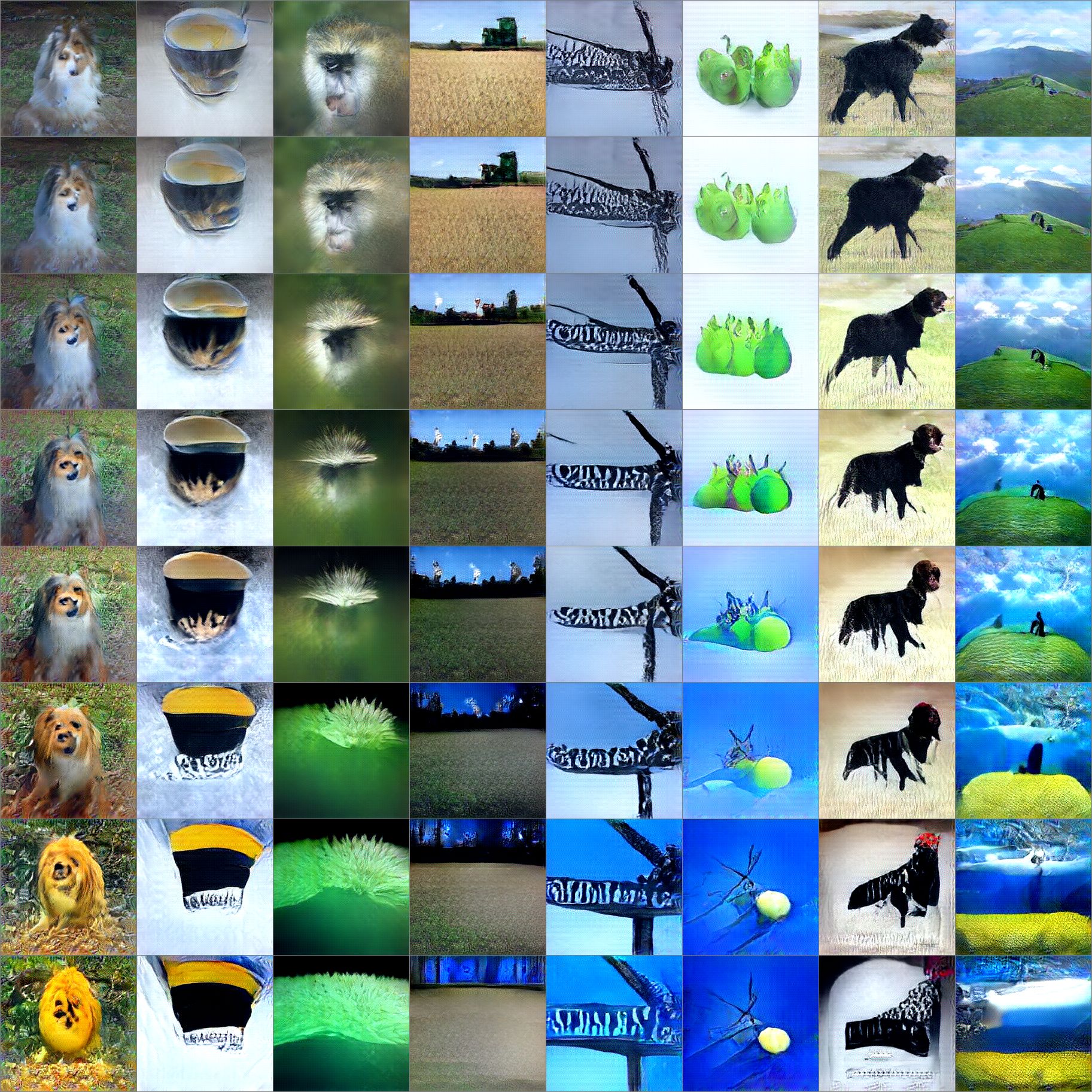} \\
\includegraphics[width=0.48\linewidth]{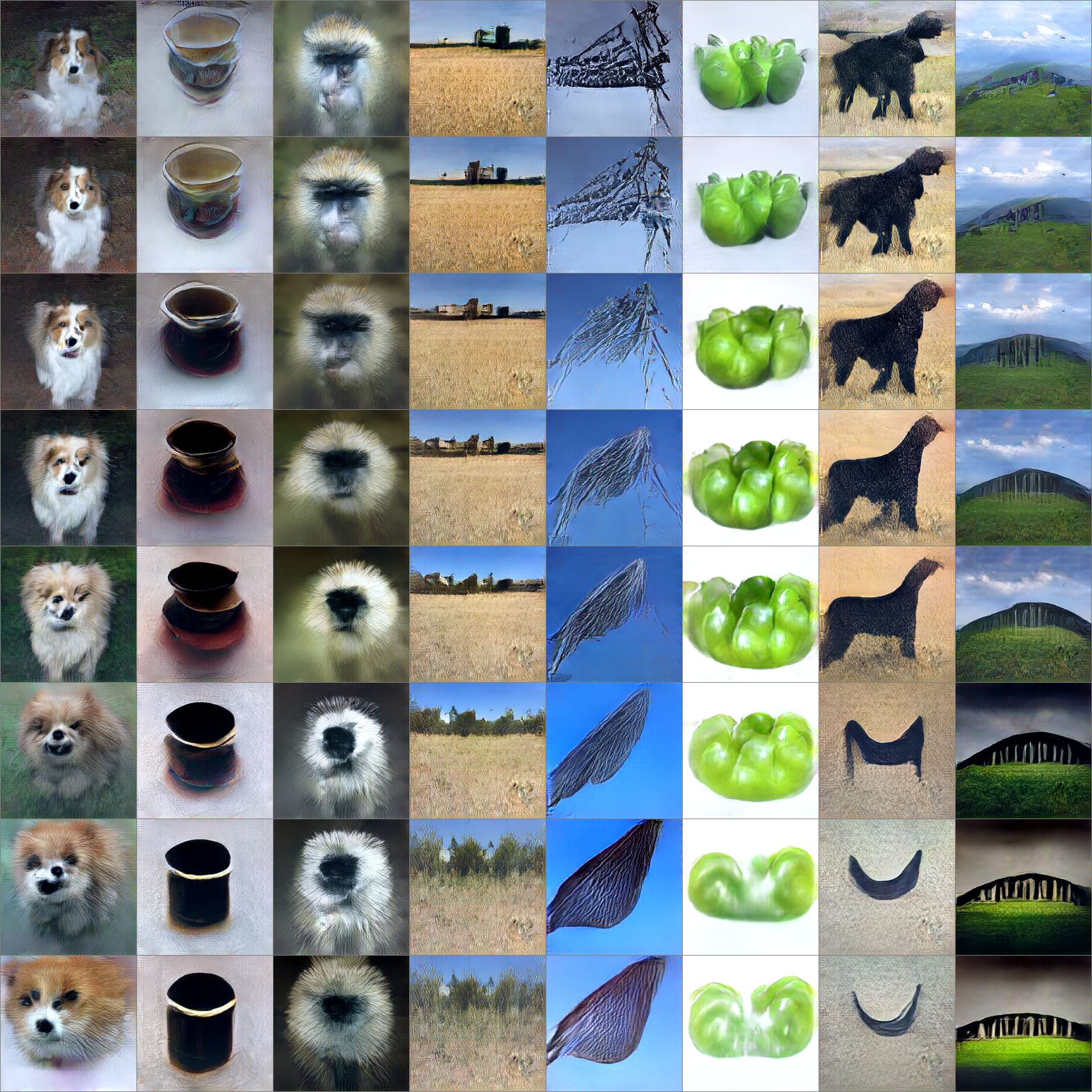} &
\includegraphics[width=0.48\linewidth]{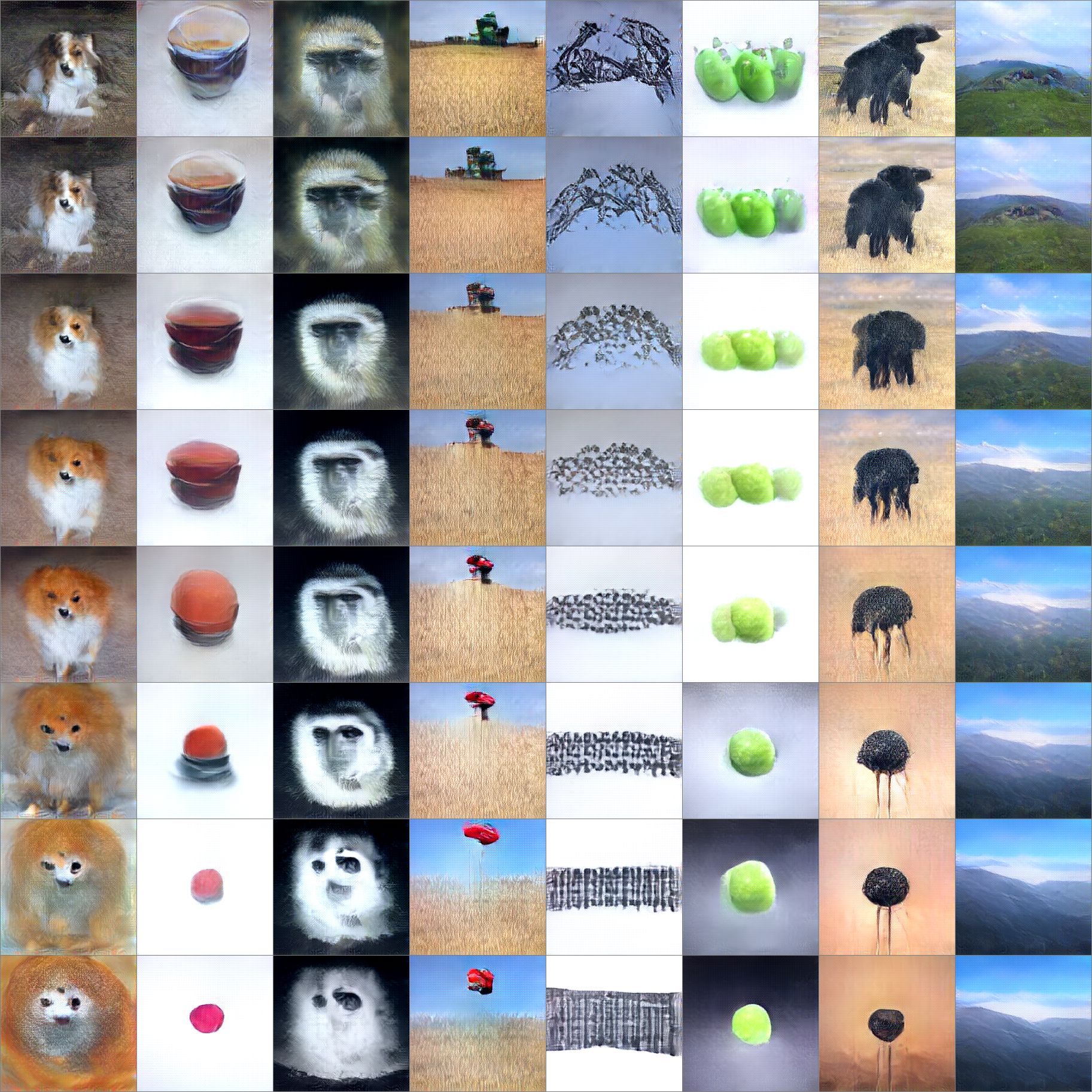}
\end{tabular}
\end{center}
\caption{Iterative re-encoding and reconstructions for different layers of AlexNet.
Each row of each block corresponds to an iteration number: 1, 2, 4, 6, 8, 12, 16, 20.
\textbf{Topmost:} input images,
\textbf{Top left:} \conv5,
\textbf{Top right:} \fc6,
\textbf{Bottom left:} \fc7,
\textbf{Bottom right:} \fc8.}
\label{fig:iterative}
\end{figure*}

\begin{figure*}
\begin{center}
\setlength{\tabcolsep}{0.1cm}
\begin{tabular}{c}
\includegraphics[width=0.48\linewidth]{ICML_orig_iterative_supp.jpg} \\
\includegraphics[width=0.48\linewidth]{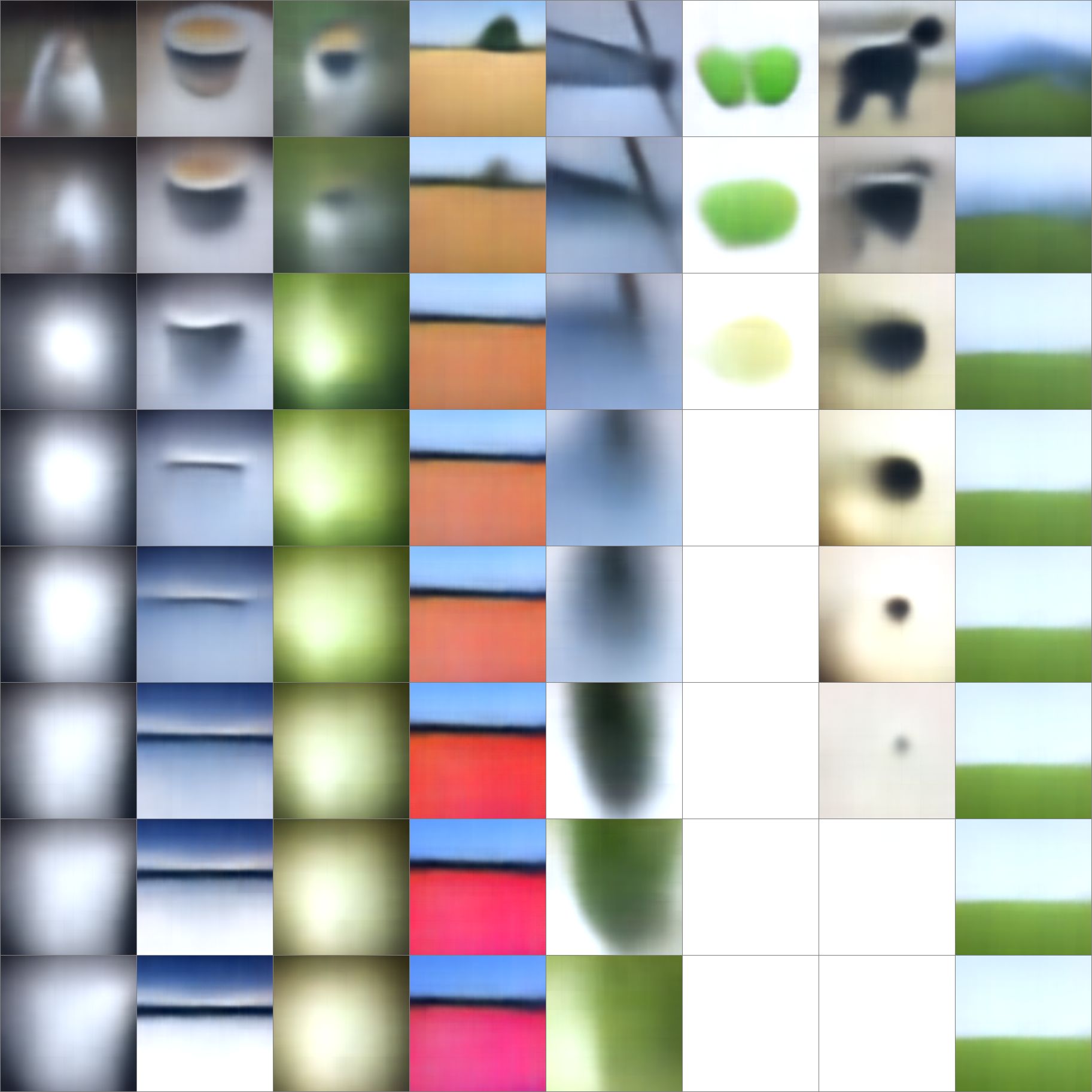}
\end{tabular}
\end{center}
\caption{Iterative re-encoding and reconstructions with network trained to reconstruct from AlexNet \fc6 layer with squared Euclidean loss in the image space.
On top the input images are shown.
Then each row corresponds to an iteration number: 1, 2, 4, 6, 8, 12, 16, 20.}
\label{fig:iterative_eucl}
\end{figure*}

Figures~\ref{fig:iterative} and~\ref{fig:iterative_eucl} show results of iteratively encoding images to a feature representation and reconstructing back to the image space.
As can be seen from Figure~\ref{fig:iterative_eucl}, the network trained with loss in the image space does not preserve the features well, resulting in reconstructions quickly diverging from the original image.

\end{document}